\documentclass{article} 
\usepackage{iclr2024_conference,times}

\usepackage{amsmath,amsfonts,bm}

\def\eqref#1{equation~\ref{#1}}

\def\1{\bm{1}}

\DeclareMathAlphabet{\mathsfit}{\encodingdefault}{\sfdefault}{m}{sl}
\SetMathAlphabet{\mathsfit}{bold}{\encodingdefault}{\sfdefault}{bx}{n}

\usepackage{hyperref}
\usepackage{url}
\usepackage{graphicx}
\usepackage{booktabs}
\usepackage{multicol}
\usepackage{multirow}
\usepackage{color}
\usepackage{array}
\usepackage{subcaption}
\definecolor{pink}{RGB}{255,153,204}

\title{A Spark of Vision-Language Intelligence: \\2-Dimensional Autoregressive Transformer for Efficient Finegrained Image Generation}

\author{Liang Chen\textsuperscript{1}, Sinan Tan\textsuperscript{2}, Zefan Cai\textsuperscript{3}, Weichu Xie\textsuperscript{4}, Haozhe Zhao\textsuperscript{1}, Yichi Zhang\textsuperscript{1} \\
\textbf{Junyang Lin\textsuperscript{2}, Jinze Bai\textsuperscript{2}, Tianyu Liu\textsuperscript{2}, Baobao Chang\textsuperscript{1}} \\
\textsuperscript{1}Peking University\quad
\textsuperscript{2}Alibaba Group\quad 
\textsuperscript{3}University of Wisconsin–Madison\\
\textsuperscript{4}Beijing Institute of Technology \quad\quad
\texttt{leo.liang.chen@outlook.com}
}

\iclrfinalcopy 
\begin{document}

\maketitle

\begin{figure}[h]
\vspace{-25pt}
\centering
\includegraphics[width=0.95\textwidth]{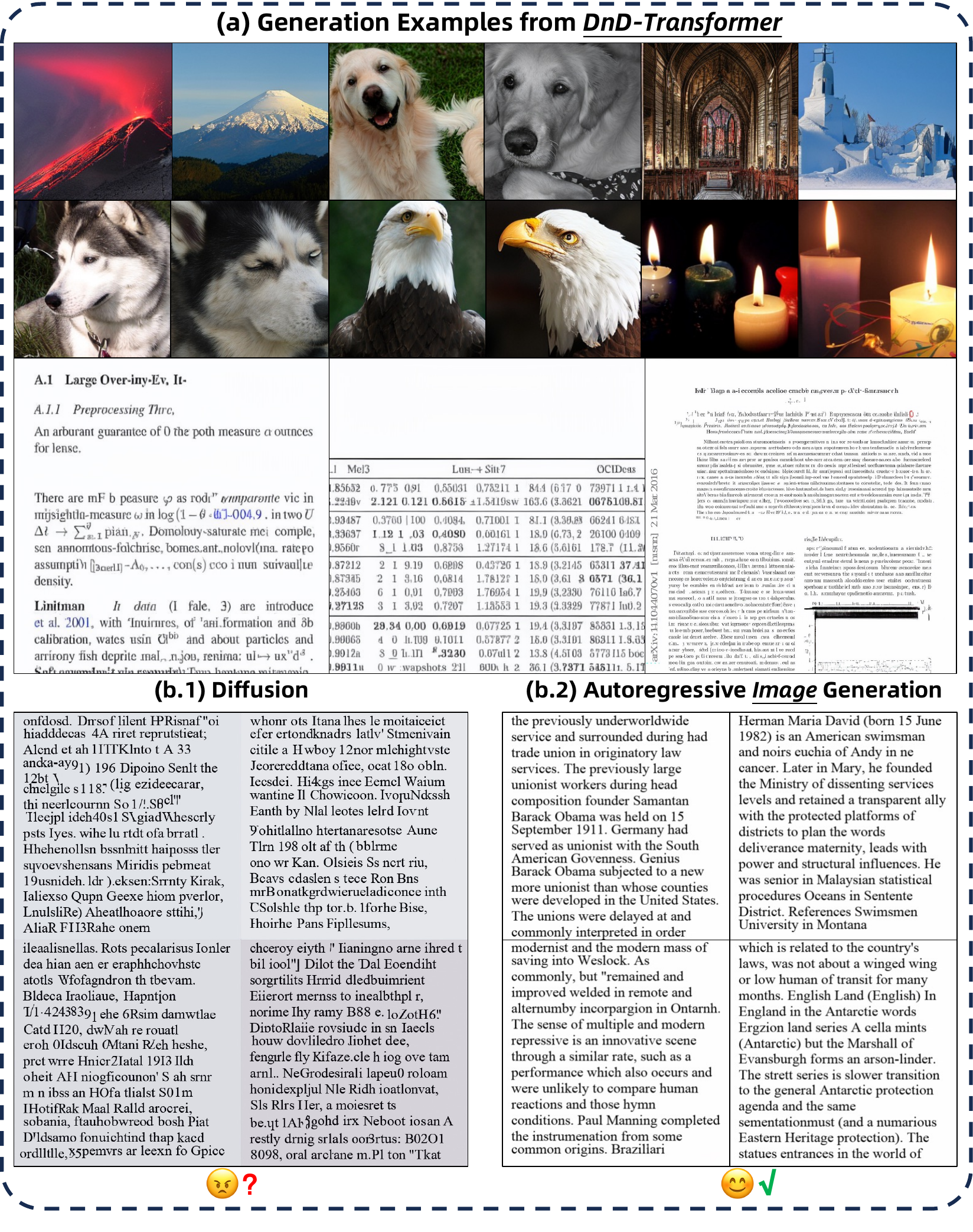}
\vspace{-15pt}
\caption{Generations from DnD-Transformers trained on class-conditional ImageNet256$\times$256 (a.top) and unconditional arXiv images (a.bottom). Unconditional rich-text image generations by trained diffusion (b.1) and autoregressive model (b.2), where autoregressive model has dominating performance, showing a spark of vision-language intelligence after purely training on images.}

\label{fig:dnd-teaser}
\end{figure}



\newpage
\begin{abstract}


This work tackles the information loss bottleneck of vector-quantization (VQ) autoregressive image generation by introducing a novel model architecture called the 2-Dimensional Autoregression (DnD) Transformer. The DnD-Transformer predicts more codes for an image by introducing a new autoregression direction, \textit{model depth}, along with the sequence length direction. Compared to traditional 1D autoregression and previous work utilizing similar 2D image decomposition such as RQ-Transformer, the DnD-Transformer is an end-to-end model that can generate higher quality images with the same backbone model size and sequence length, opening a new optimization perspective for autoregressive image generation. Furthermore, our experiments reveal that the DnD-Transformer's potential extends beyond generating natural images. It can even generate images with rich text and graphical elements in a self-supervised manner, demonstrating an understanding of these combined modalities. This has not been previously demonstrated for popular vision generative models such as diffusion models, showing a spark of vision-language intelligence when trained solely on images. Code, datasets and models are open at \href{https://github.com/chenllliang/DnD-Transformer}{https://github.com/chenllliang/DnD-Transformer}.


\end{abstract}

\section{Introduction}

The field of autoregressive (AR) image generation is experiencing a resurgence of interest, largely driven by groundbreaking advancements in large language models (LLMs), exemplified by the release of ChatGPT \citep{openai2022chatgpt}. Because typical AR image generation methods also predict output in a next-token prediction manner, this resemblance has sparked significant efforts in two main areas: 1) transferring advanced, large-scale training techniques and expertise from LLMs to AR image generation models \citep{lvm,var,llamagen}, and 2) developing truly multimodal foundation models capable of both understanding and generating multimodal information within a unified training framework \citep{unified-io,unified-io2,chameleon}. These developments have the potential to lead to more versatile and powerful multimodal AI systems.

A review of the development history of AR image generation approaches reveals significant efforts focused on finding better sequential decompositions of images and balancing reconstruction fidelity with prediction difficulty. Early models, like PixelCNN \citep{pixelcnn}, generated images pixel by pixel. This approach was later enhanced by using vector-quantized variational autoencoders (VQVAEs) to compress images and model the prior distribution of discrete tokens in a compact latent space \citep{vqvae}. Vector quantization (VQ) paved the way for notable models such as VQGAN \citep{vqgan}, DALL·E \citep{dalle1}, and MUSE \citep{muse}, and it remains a core technique in recent AR image generation models like VAR \citep{var} and LlamaGen \citep{llamagen}, and multimodal foundation models like LVM \citep{lvm}, Unified-IO \citep{unified-io,unified-io2}, and Chameleon \citep{chameleon}.

However, despite advancements in AR image generation, VQ-based autoregressive methods face two persistent criticisms, especially juxtaposed with latent diffusion models~\citep{ldm}:

\begin{figure}[t]
\centering
\includegraphics[width=0.9\textwidth]{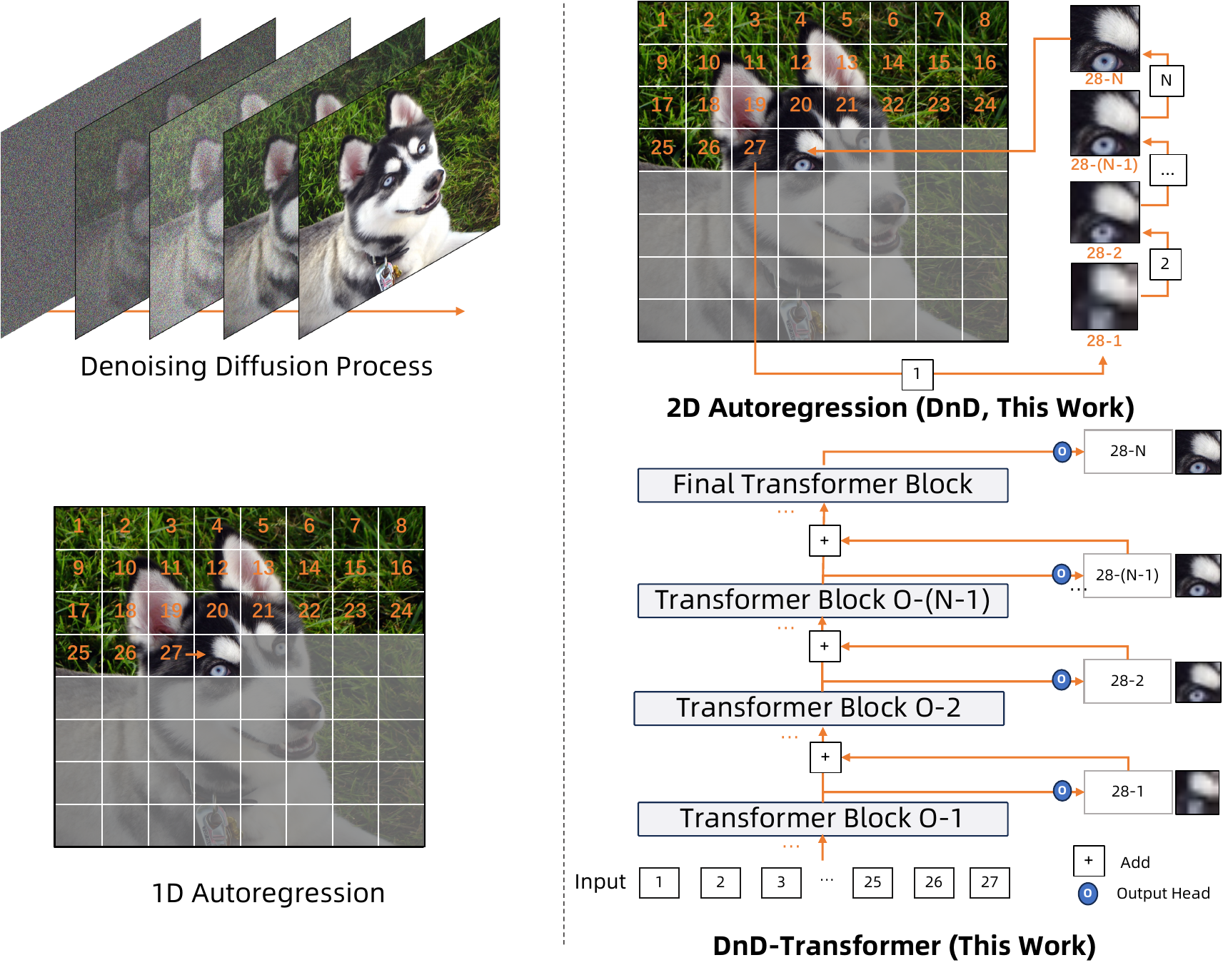}
\caption{Illustration of the proposed DnD-Transformer. N denotes the number of depth autoregression. O-i denotes the transformer layer index for the i-th prediction head. Each transformer layer predicts the corresponding depth code, achieving multi-code prediction within one forward pass.}
\label{fig:dnd-front}
\vspace{-5mm}
\end{figure}


\paragraph{1) Information loss inherent in the quantization process.} Quantization, specifically in VQVAE, introduces significant information loss. With a typical configuration (N=8192, f=16), the Information Compression Ratio ($\text{ICR}=\frac{\log N}{24f^2}$, explained in Equation~\ref{eq:ICR}) is just 0.21\%, drastically lower than the 8.3\% of Stable Diffusion's VAE\footnote{The Stable Diffusion VAE (\url{https://huggingface.co/stabilityai/sd-vae-ft-mse}) uses a downsampling factor (f) of 8 and 4 channels, with fp32 tensor precision ($\log N=4\times32$).}, hindering fine-grained detail reconstruction. According to Chameleon \citep{chameleon}, the authors note that their VQ tokenizer struggles to reconstruct finegrained details like text in images, which we believe is due to the low ICR of their tokenizer.

\paragraph{2) Substantially increased computational requirements for producing higher-quality images.}  According to Equation~\ref{eq:ICR},Increasing ICR by expanding the latent space (N) is logarithmically limited and computationally expensive leading to potential codebook collapse and more embedding parameters, while reducing the downscaling factor (f) significantly increases computational overhead due to a longer token sequence of $O(1/f^{2})$ and a higher transformer computation complexity of $O(1/f^{4})$.

We draw inspiration from the Residual Quantization method~\citep{rqvae}, which provides a new dimension for sequentially decomposing the image for better generation quality. However, the proposed RQ-Transformer employs two separate transformer models. This structure presents difficulties in integrating current LLMs for end-to-end training. In this work, we aim to solve the problem covering the two mentioned concerns: \textbf{\textit{Can we overcome the information loss  of VQ-based AR image generation without increasing overall computation budget in an end-to-end manner?}} 

We propose a novel paradigm for AR image generation called 2-Dimensional Autoregression (DnD) and DnD-Transformer, an end-to-end model architecture. DnD Autoregression introduces a new depth dimension along with the original spatial dimension. In the depth dimension, the image patch could be decomposed in any causal coarse-to-fine order, including the residual decomposition~\citep{rqvae}, Gaussian denoising decomposition~\citep{ddpm} and etc. With a depth of $d$ and other configurations unchanged, the ICR of DnD Autoregression becomes $d\times\frac{\log N}{24f^2}$, more effectively reducing the information loss comparing to increasing the codebook size $N$.

The remaining problem is how to predict the $d$ times more tokens effectively. We propose the DnD-Transformer. As shown in Figure~\ref{fig:dnd-front}, it inserts multiple prediction heads into the backbone transformer decoder model to predict the depth codes and conduct additional autoregressive predictions in each forward process. Different from RQ-Transformer~\citep{rqvae}, the DnD-Transformer does not require additional modules or increased sequence length, making it applicable to any language model architecture and efficiently generate more fine-grained images.

Our experiments show several interesting results: 

\begin{enumerate}
    \item Superior reconstruction of fine-grained image details using residual image decomposition in VQVAEs, disproving VQ's limitations with text-rich images
    \item More efficient and lower-entropy decomposition with DnD autoregression compared to 1D methods, evidenced by lower training cross-entropy loss despite predicting more codes
    \item Significant outperformance of the AR baseline on ImageNet 256x256 generation, achieving up to 1.54 FID and 82.6 IS improvements (XXL model, cfg=2) without increased model size or sequence length, even surpassing larger LlamaGen model trained with longer sequence length
    \item \textbf{A spark of vision-language intelligence} for the first time, enabling unconditional rich-text image generation, outperforming diffusion models like DDPM and Stable Diffusion on dedicated rich-text image datasets, highlighting the distinct advantage of autoregressive models for multimodal modeling.
\end{enumerate}

\section{2D Visual Tokenizer and 2D Autoregression}

\subsection{Understand VQVAE as Compression}

We introduce the basics of AR generation in Section~\ref{app:vq} in the appendix. We can better understand the reconstruction ability of VQVAE from the lens of compression.  Let us assume a VQVAE with down-scaling factor $f$, codebook size $N$, input image's size of $H\times W$, then the shape of the quantized code is $h \times w = (H/f) \times (W/f)$. We assume that the code follows a uniform distribution, so each code has $\log N$ bits information. Its information compression ratio (ICR) is as follows.

\begin{equation}
    ICR(N,f) = \frac{(H/f) \times (W/f) \times \log N}{H\times W\times3 \times \log 256} = \frac{\log N}{24 f^2}
    \label{eq:ICR}
\end{equation}

A typical configuration (N=8192, f=16) results in 0.21\% ICR. This ICR is significantly lower than JPEG's 5\% ICR~\citep{jpeg_compression_rate}. To increase ICR, the 1D AR method could increase $N$ (might face the codebook collapse problem~\citep{fsq} and the improvement is logarithmically bounded) or decrease $f$ (more effective, but increases the token count quadratically).

\subsection{Images' 2D Decomposition and Quantization}

As pointed out by Equation~\ref{eq:ICR}, the information compression ratio of VQVAE is bounded by the size of the codebook and the downscaling ratio. Residual Quantization~\citep{rqvae} proposes a new direction to quantize the image feature with multiple residual codes to reduce the quantization error and improve the quality of the reconstruction. For a feature map having $h\times w$ vectors, RQVAE uses $h\times w \times d$ codes to quantize the feature map, where $d$ is the depth dimension of the code. For each feature vector $\mathbf{v}$, RQ finds $d$ codes ($q_1, q_2, ..., q_d$) by sequentially conducting $d$ times residual decomposition and quantization operation $\mathcal{Q}(x)$ as finding the closest entry to $x$ from the codebook:

\begin{equation}
\begin{aligned}
q_d = \mathcal{Q}(r_{d-1}), \quad r_d = r_{d-1} - q_d, \quad r_0 = \mathbf{v}
\end{aligned}
\label{eq:rq}
\end{equation}

Consequently, the sum of the residual codes $\sum_{i=1}^{d} q_{i}$ is expected to approximate more closely the feature vector $\mathbf{v}$, thus reducing the quantization error. We generalize this process as two-dimensional autoregression (DnD), which extends beyond Markov residual decomposition and can be applied to any decomposition operation, such as the diffusion process~\citep{ddpm}, etc. 

\begin{figure}[t]
\centering
\includegraphics[width=\textwidth]{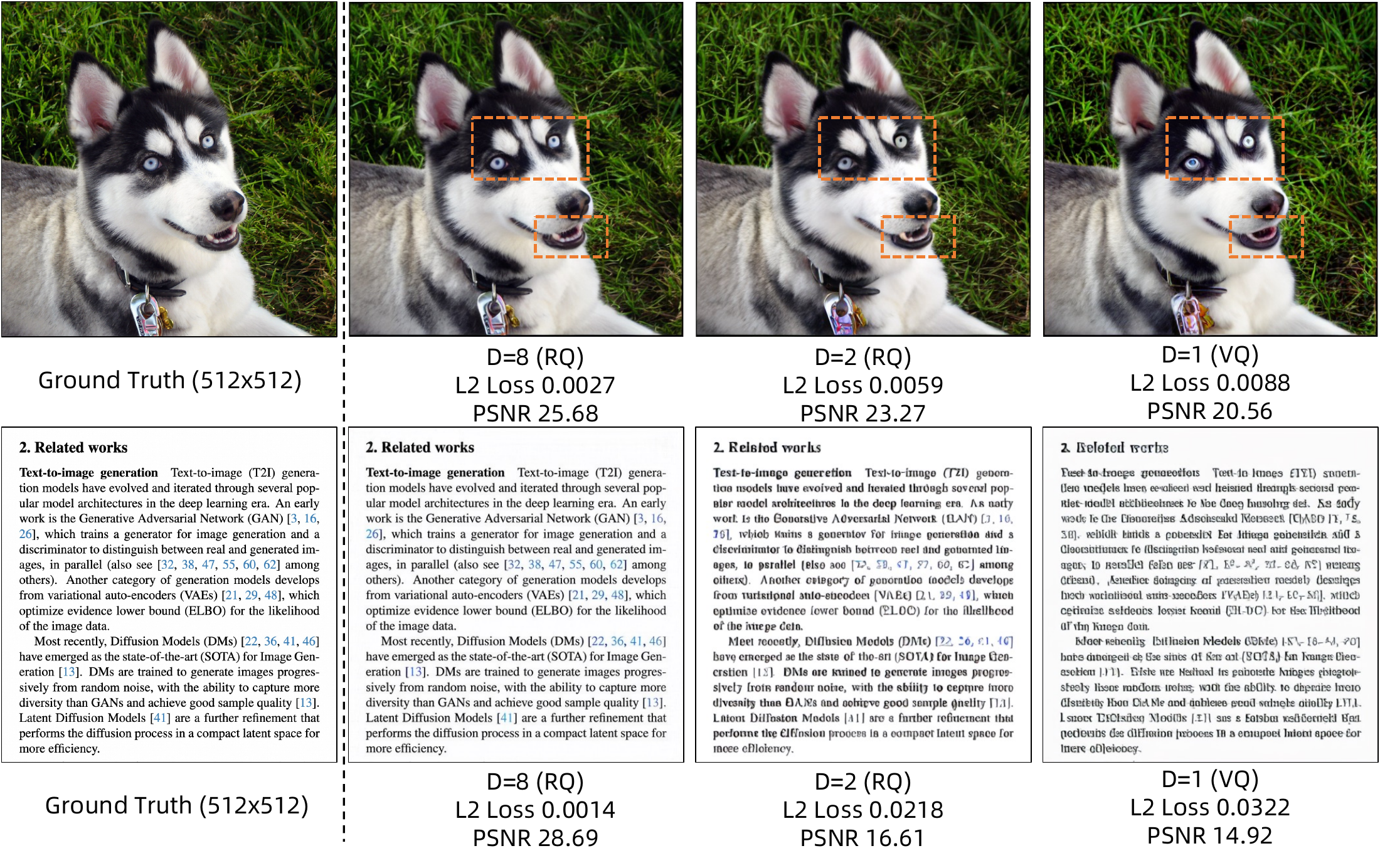}
\caption{Performance of our visual tokenizers of different depths. The reconstruction of complex features (i.e., eyes, mouse and text) gains significant improvement as the depth increases.}
\label{fig:reconstruct_im_text}
\end{figure}

DnD Autoregression quantizes a 2D feature map $\mathbf{m} \in \mathcal{R}^{h\cdot w\cdot c}$ by decomposing it in two directions. First, $\mathbf{m}$ is divided into $h\cdot w$ feature vectors.  Second, each vector $\mathbf{v}$ is decomposed into $n$ codes $(q_1, ..., q_n)$ using a function $\mathcal{D}^n(\mathbf{v},\mathcal{Q})$ based on a codebook $\mathcal{Q}$. The resulting quantized map $\mathbf{q}$ has shape $h\cdot w\cdot n$ and is predicted in depth-first-spatial-second order. This decomposition could also be non-Markov, unlike RQVAE. The selection of potentially better decomposition functions is left for future exploration. We still use the residual quantization from Equation~\ref{eq:rq} as $\mathcal{D}^n$.  DnD decomposition increases the ICR $d$ times (Equation~\ref{eq:ICR_dnd}), more effectively than increasing codebook size.  The remaining challenge of predicting $d$ times more codes is addressed by our DnD-Transformer.

\begin{equation}
    ICR(N,f,d) = d\times \frac{(H/f) \times (W/f) \times \log N}{H\times W\times3 \times \log 256} = d\times\frac{\log N}{24 f^2}
    \label{eq:ICR_dnd}
\end{equation}

\subsection{Reconstruction Performance}



We evaluate the reconstruction performance of our trained visual tokenizers with varying maximum codebook depths using the standard ImageNet dataset as the benchmark. All images are resized to 256×256 resolution. We train the different visual tokenizers using the same training objectives as in ~\citet{rqvae}, and assess the reconstruction Fréchet Inception Distance (rFID) on the ImageNet validation set using ADM's evaluation suite ~\citep{adm}. The results are presented in Table~\ref{tab:rfid}. For comparison, we include the rFID from the VAE of SDXL~\citep{sdxl} and Stable-Diffusion 3~\citep{sd3} . Our findings demonstrate that our trained visual tokenizer achieves an rFID lower than 1 with two or more codebook depths, even surpassing the performance of SD3's continuous VAE with less theoretical information loss. As shown in the example from  Figure~\ref{fig:reconstruct_im_text}, by increasing code depth, we could reconstruct more fine-grained details in the image.

\begin{table}[t]  
\centering  
\begin{subtable}{.48\textwidth} 
\centering  
\small 
\setlength{\tabcolsep}{3pt} 
\begin{tabular}{c|ccc}  
\toprule  
\multirow{2}{*}{\begin{tabular}[c]{@{}c@{}}Depth\\\end{tabular}} & \multicolumn{3}{c}{ImageNet 256$\times$256} \\
\cmidrule(lr){2-4}  
& rFID$\downarrow$ & L2 Loss$\downarrow$ & Code Usage$\uparrow$  \\
\midrule  
1 & 2.98 & 0.11 & 100\%  \\
2 & 0.93 & 0.08 & 100\%  \\
4 & 0.60 & 0.05 & 100\%  \\
8 & 0.42 & 0.04 & 100\%  \\
\midrule  
SDXL & 0.68 & 0.05 & - \\
SD3 & 0.67 & 0.04 & -  \\
\bottomrule  
\end{tabular}  
\caption{\textbf{Reconstruction Performance on ImageNet 256$\times$256 Validation Set.}}  
\label{tab:rfid}  
\end{subtable}  
\hfill  
\begin{subtable}{.48\textwidth} 
\centering  
\small 
\setlength{\tabcolsep}{3pt} 
\begin{tabular}{c|ccc}  
\toprule  
\multirow{2}{*}{\begin{tabular}[c]{@{}c@{}}Depth\\\end{tabular}} & Text256 & Text512 & arXiv512 \\   
\cmidrule(lr){2-4}  
& \multicolumn{3}{c}{rOCR$\uparrow$} \\   
\midrule  
1 & 0.15 & 0.73  & 0.14 \\
1$^\dagger$ & 0.00 & 0.00  & 0.00 \\
2 & 0.50 &  0.81 & 0.49 \\
8 & 0.80 &  0.83 & 0.67 \\
\midrule  
SDXL & 0.72 & 0.83 & 0.66\\
SD3 & 0.82 & 0.83 & 0.74 \\
\bottomrule  
\end{tabular}  
\caption{\textbf{Reconstruction OCR Performance.} $\dagger$ indicates zero-shot tokenizer trained on ImageNet.}  
\label{tab:rocr}  
\end{subtable}   
\caption{\textbf{Ablation studies on the reconstruction performance of visual tokenizers}. Our trained tokenizers all have a $f=16$ downscaling factor and $N=16384$ codebook size.}  
\label{tab:reconstruction}  

\end{table}


\begin{figure}[t]
    \centering
    \begin{subfigure}[b]{0.48\textwidth}
        \centering
        \includegraphics[width=\textwidth]{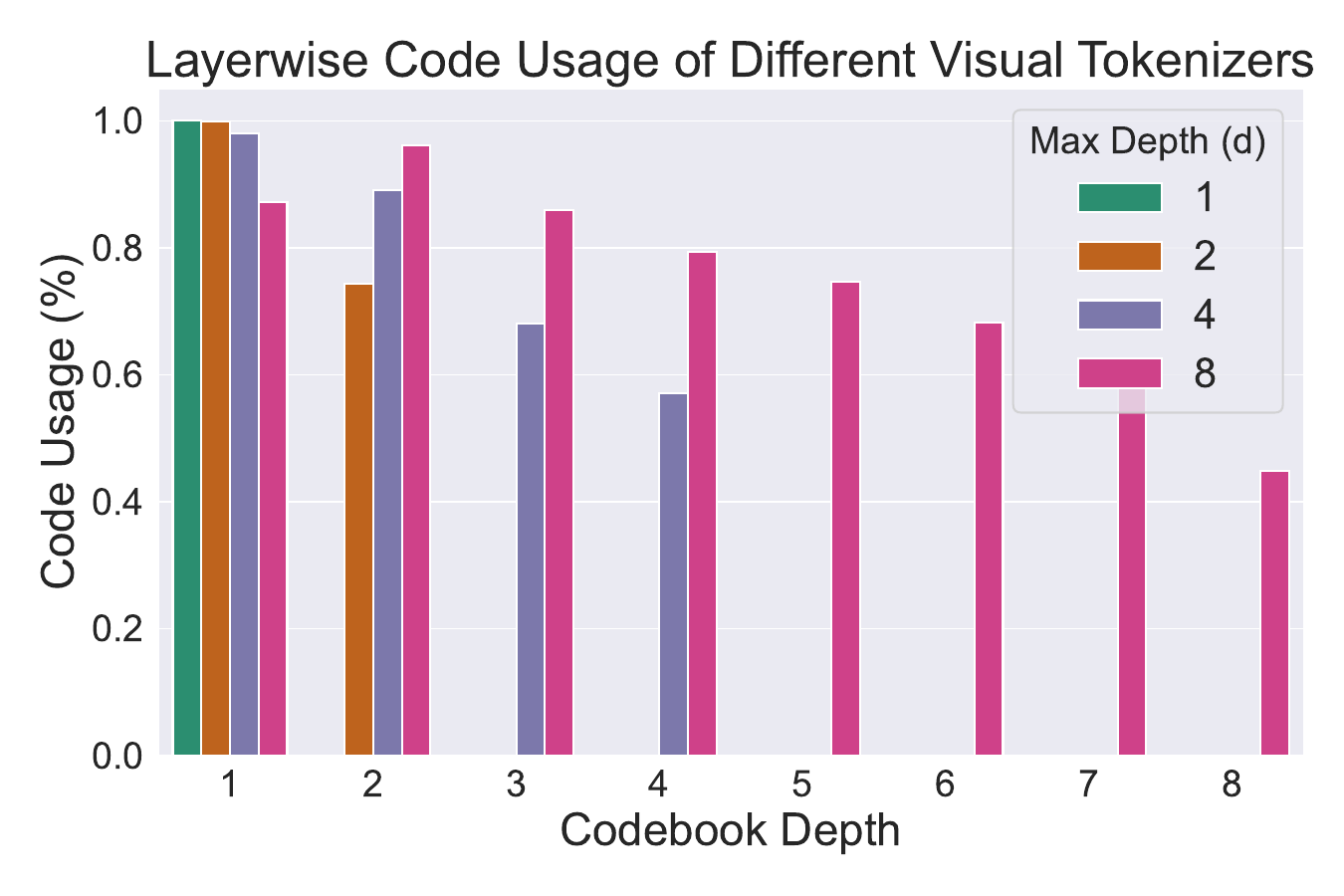}
        \caption{Layerwise code usage of visual tokenizers.}
        \label{fig:code_usage}
    \end{subfigure}
    \hfill
    \begin{subfigure}[b]{0.48\textwidth}
        \centering
        \includegraphics[width=\textwidth]{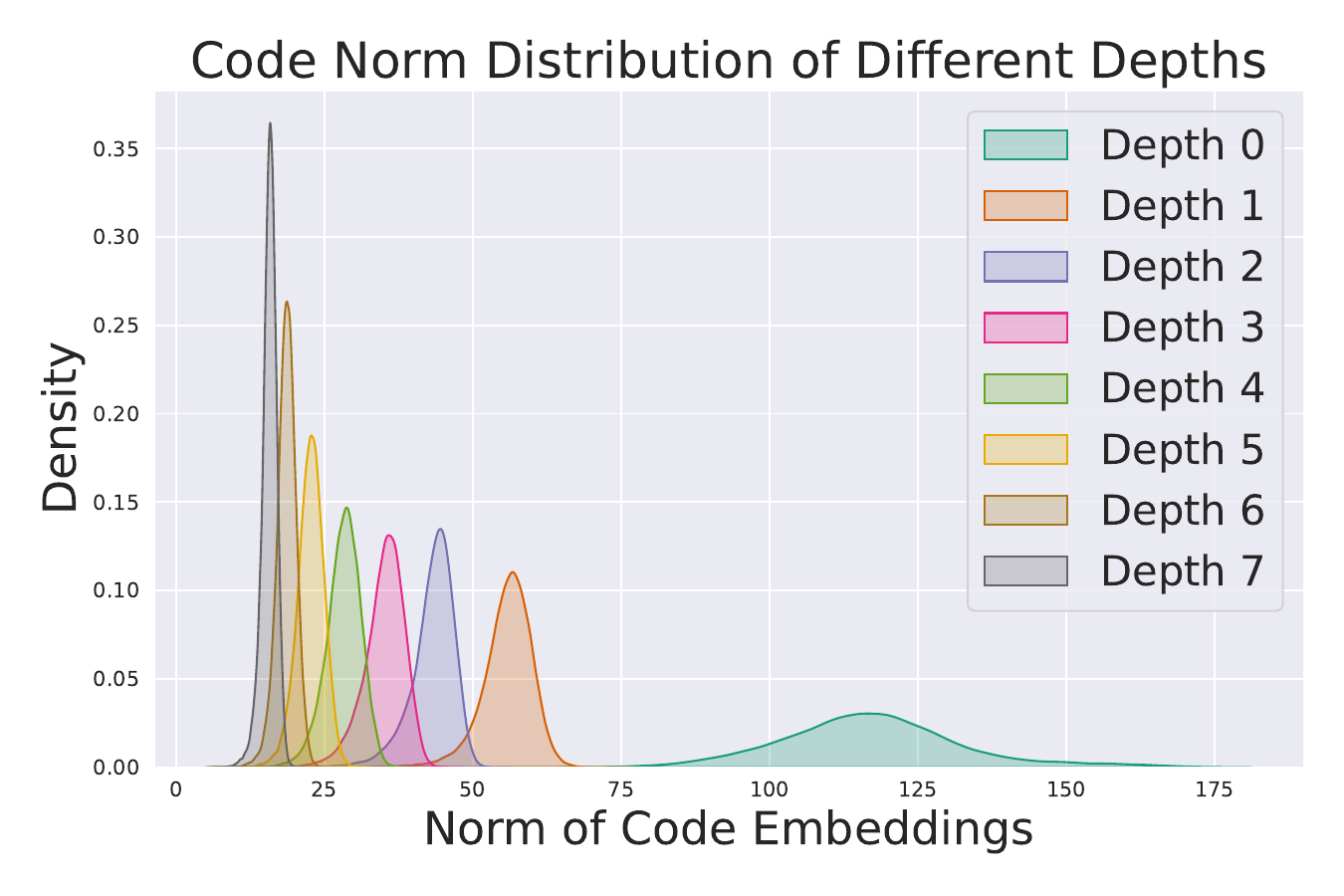}
        \caption{Code Norm Distribution for Tokeniers }
        \label{fig:density}
    \end{subfigure}
    \caption{Analysis of visual tokenizers.}
    \label{fig:combined}
    \vspace{-5mm}
\end{figure}




\paragraph{Code Usage.}  We further analyze the code usage in each codebook layer, with results shown in Figure~\ref{fig:code_usage}. The analysis indicates that usage generally decreases as depth increases. This is due to the diminishing diversity of code usage as the residual decomposition progresses deeper, resulting in smaller feature norms and more centralized code usage according to Figure~\ref{fig:density}. Interestingly, we do not observe signs of codebook collapse with the DnD visual tokenizers, even when using a large codebook size (16384), as mentioned in previous work~\citep{fsq}. While they reported much lower code usage ($<50\%$), our tokenizer achieves $100\%$ usage across all maximum depths.

\subsection{VQVAEs Can Perfectly Reconstruct Rich-text Images}

A prevalent criticism of VQVAE has been its alleged intrinsic information loss problem, particularly its inability to reconstruct images with fine details, such as those containing rich text~\citep{chameleon}. However, we argue that this claim is unfounded. Our findings suggest that VQVAE can indeed achieve perfect reconstruction of detailed images, when provided sufficient data and an increased number of codes used to represent each image. This demonstrates that the perceived limitations of VQVAE can be overcome through appropriate data-centric adjustments and model scaling-up.


\paragraph{rOCR - A New Metric.} We proposes rOCR, a novel metric for evaluating rich-text image reconstruction. Unlike rFID/L2 Loss, rOCR measures textual recognizability using the Qwen2-VL-72B~\citep{wang2024qwen2vlenhancingvisionlanguagemodels} visual language model for OCR. The metric computes the Rouge-L score between recognized and groundtruth text (or original image OCR if groundtruth is unavailable).


\paragraph{Experiments and Results.}


Two rich-text image datasets, Text-Image and arXiv-Image (details in Section~\ref{sec: task and datasets}), were used to train visual tokenizers. Performance (rOCR scores) was evaluated on both datasets' 1K test sets, compared against ImageNet-trained tokenizers, SDXL's~\citep{sdxl} and Stable-Diffusion-3's VAE~\citep{sd3}. Text-Image was also tested at a reduced 256$\times$256 resolution to assess resolution impacts. Table~\ref{tab:rfid} shows the rOCR results, with reconstruction examples in Figures~\ref{fig:reconstruct_im_text} and~\ref{fig:reconstruct_text_ablation}. Results indicate more training data and deeper tokenizers improve text reconstruction. Unlike ~\citet{chameleon}, our discrete visual tokenizers excel in rich-text image reconstruction even compared to continuous VAEs.




\section{The DnD-Transformer}

Prior section showed DnD visual tokenizers effectively reconstruct fine details like text. However, efficiently predicting the increased number of depth codes ($d$ times more) remains challenging. Existing methods, like RQ-Transformer, use a separate transformer for depth, hindering integration with LLMs. We propose an efficient end-to-end architecture for multi-code prediction.

\subsection{DnD-Transformer Design}

\begin{figure}[t]
\centering
\includegraphics[width=\textwidth]{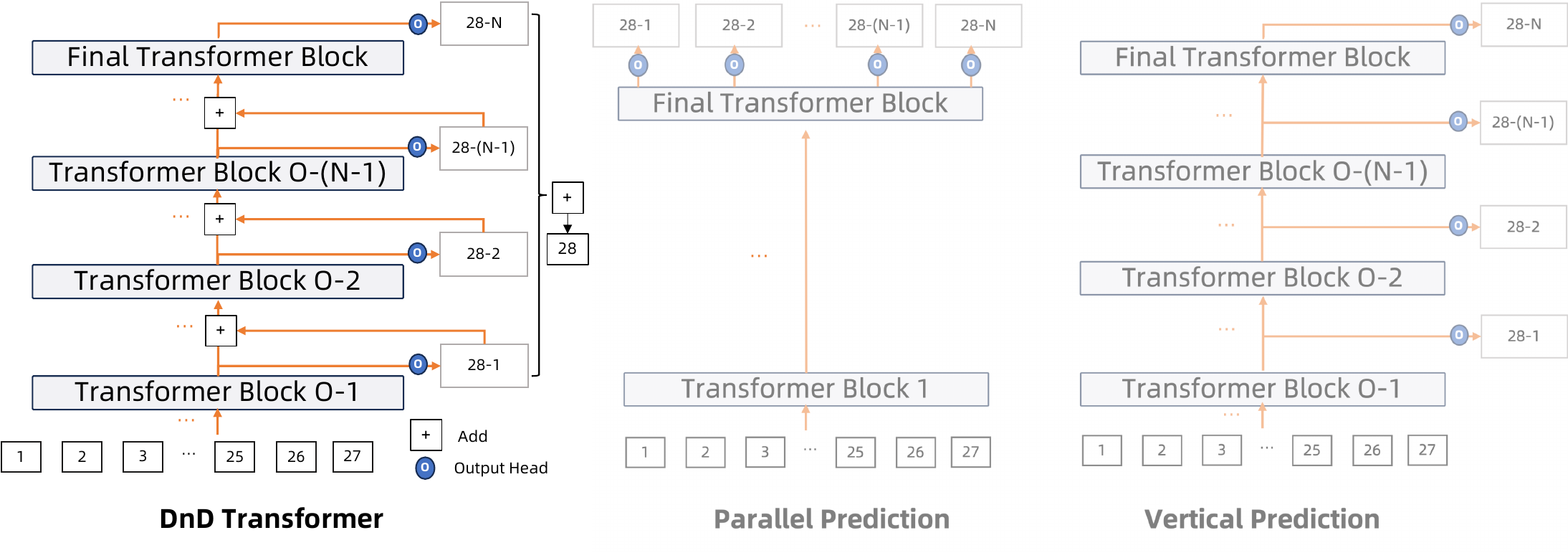}
\caption{Different explored multi-token prediction architectures for DnD-Transformer, which are all designed to generate multiple codes with one forward pass.}
\label{fig:methods}
\vspace{-5mm}
\end{figure}

Figure~\ref{fig:methods} shows DnD-Transformer and its variants: Parallel and Vertical Prediction. Parallel Prediction adds multiple prediction heads for simultaneous multi-depth code prediction, similar to accelerated LLM inference~\citep{cai2024medusasimplellminference}. However, this ignores the coarse-to-fine nature (Figure~\ref{fig:density}) of code distributions, where deeper codes have smaller norms and are more centered. Vertical Prediction addresses this by sequentially predicting codes. Adding autoregression further refines this by conditioning deeper code predictions on previous ones, achieving the best multi-layer code prediction without increasing model parameters or sequence length. Ablation on the structure design is shown in Table~\ref{tab:abaltion} from Appendix.

\subsection{Implementation Details}

As shown in the left part of Figure~\ref{fig:methods}, the increment of DnD-Transformer compared to vanilla transformer decoder is the additional output head and embedding add operation. Let's assume the linearized codemap's length is $L=h\times w$ and code depth is $d$. During generation, DnD-Transformer conducts $L$ forward process and each forward process generate $d$ codes sequentially. After generating codes for all depths in a forward process, the embeddings of all codes are added up as the next input token. In this way, the model could generate $L\times d$ tokens with only $L$ forward passes, improving the generation quality with the same inference cost as standard 1D auto-regression transformer. The only additional hyper-parameter is the layer indexes to predict code of different depths. We adopt the same transformer decoder's architecture as LLaMA~\citep{llama1} and , please refer to Appendix~\ref{app:dnd-transformer} for the training details of our DnD-Transformer.

\section{Experiments and Findings}
\subsection{Tasks and Datasets}
\label{sec: task and datasets}
\paragraph{Class-Conditional Image Generation.} We conduct standard conditional image generation task with ImageNet-1k benchmark. Images are resized to 256$\times$256 resolution during training and evaluation. We sample 50k images with classes uniformly distributed, and compute the FID, IS, Precision and Recall aganist the training set data using the ADM evaluation tool~\cite{adm}.

\begin{figure}[t]
\centering
\includegraphics[width=0.6\textwidth]{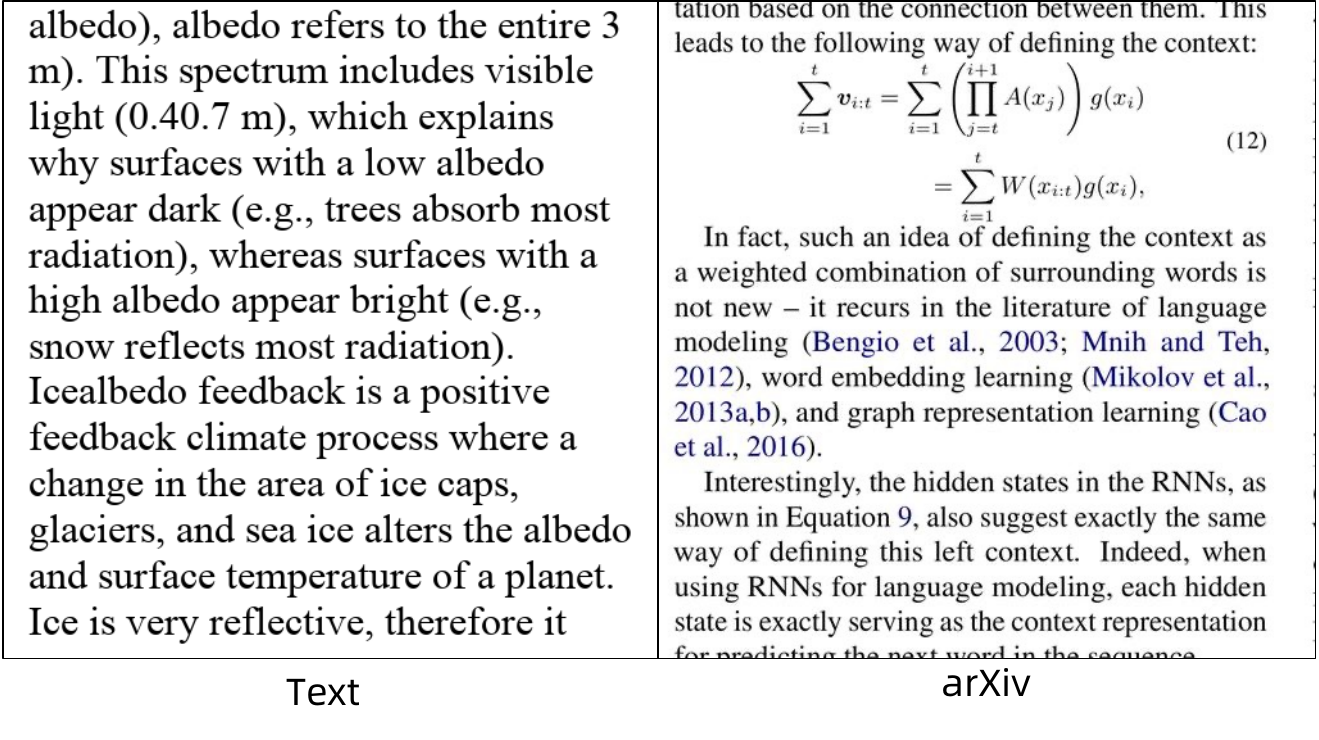}
\caption{Data examples in of the collected Text-Image and arXiv-Image image datasets. }
\label{fig:data-example}
\vspace{-5mm}
\end{figure}

\paragraph{Unconditional Rich-Text Image Generation.} We collect two datasets for this task. Dataset examples are shown in Figure~\ref{fig:data-example}. Models are trained in a unconditional setting in this task. We aim to explore whether the tested vision generation models could understand and generate the complex logical interrelation among the generated elements such as language. 

\begin{enumerate}
    \item \textit{Pure Text Images (Text-Image).} The dataset is automatically rendered from a portion of English wikipedia~\citep{wikidump}, consisting of 2.4M images. Each image has a original resolution of 512$\times$512 and a font size of 32pt. We set a maximum of 100 words in each image with a paddling margin of 20pt. We use the \href{https://github.com/python-pillow/Pillow}{\textsc{Pillow}} library to render the image. 
    \item \textit{arXiv Images (arXiv-Image)} we first download the papers in PDF format from \url{arXiv.org}, and render the pages to image of A4 resolution ($1260\times1782$) with \href{https://github.com/Belval/pdf2image}{\textsc{pdf2image}} tool. We then randomly crop ten 512$\times$512 image from each pages and finally collect 2M images.
\end{enumerate}

We have developed an evaluation pipeline that combines Optical Character Recognition (OCR) and Perplexity Measurement for assessing the quality of generated images, with a focus on the textual information they contain. Initially, we employ the state-of-the-art open-source Vision-Language Model, Qwen2-VL-72B, to extract text from the generated images. Subsequently, we utilize the Qwen2.5-72B model to calculate the perplexity of the generated text, where the LLM is regraded as the evaluator. The resulted score is called $PPL_{ocr}$, we also test the score of groundtruth data from the training images as the performance upper-bound.

\subsection{Models}
\paragraph{Visual Tokenizers.} We train our visual tokenizer based on RQVAEs~\citep{rqvae}. We train tokenizers with code depths of $\{1,2,4,8\}$ and scaling factor $f=16$ across different experiments. We choose the checkpoint with best rFID across 150 epochs. Performance comparison of different visual tokenizers is shown in Table~\ref{tab:reconstruction}. We follow ~\citet{rqvae} to train the visual tokenizers. Details of the training of visual tokenizers are listed in Appendix~\ref{app:training-tokenizer}. Reconstruction performance of the trained visual tokenizers is shown in Table~\ref{tab:reconstruction}.

\paragraph{DnD-Transformer.} We train two size of DnD-Transformers across our experiment, namely DnD-Transformer-XXL (1.4B) and DnD-Transformer-XXXL (2.5B). Basically, DnD-Transformer inherits the LLaMA~\citep{llama1} architecture. The XXL version strictly align with the LlamaGen-XXL baseline to be fairly compared. Details of the model are shown in Appendix~\ref{app:dnd-transformer}.

\paragraph{Implemented Baselines for Class-Conditional Image Generation.} LlamaGen~\citep{llamagen} is the major baseline and state-of-the-art model for AR image generation on ImageNet. Our implemented code primarily refers to the same training codebase for fair comparison. LlamaGen could be also viewed as a special version of DnD-Transformer where the decomposition depth equals to 1.

\begin{table}[t]
\vspace{-5mm}
\centering
   \resizebox{\textwidth}{!}{
\begin{tabular}{c|lc|cccc}
\toprule
Type & Model & \#Para. & FID$\downarrow$ & IS$\uparrow$ & Precision$\uparrow$ & Recall$\uparrow$  \\

\midrule
\multirow{4}{*}{Diffusion-Reported} & ADM~\citep{adm}  & 554M       & 10.94 & 101.0        & 0.69 & 0.63    \\
 & CDM~\citep{cdm}   & $-$       & 4.88  & 158.7       & $-$  & $-$   \\
 & LDM-4~\citep{ldm} & 400M     & 3.60  & 247.7       & $-$  & $-$  \\
 & DiT-XL/2~\citep{dit}  & 675M  & 2.27  & 278.2       & 0.83 & 0.57   \\

\midrule
\multirow{6}{*}{AR-Reported}  &VQGAN~\citep{vqgan}  & 1.4B  & 5.20  & 280.3  & $-$  & $-$ \\
 & RQTransformer~\citep{rq}       & 3.8B  & 7.55  & 134.0  & $-$  & $-$     \\
  & LlamaGen-XXL (cfg=2) ~\citep{llamagen}  & 1.4B & 3.64 & 296.5 & 0.86 & 0.51 \\
 & LlamaGen-XXL$^\dagger$ (384$\times$384, cfg=2) ~\citep{llamagen}  & 1.4B & 2.52 & 295.4 & 0.84 & 0.56 \\
 & LlamaGen-3B (cfg=2) ~\citep{llamagen} & 3.1B & 4.21 & 325.2 & 0.87 & 0.49 \\
 & LlamaGen-3B$^\dagger$ (384$\times$384, cfg=2) ~\citep{llamagen} & 3.1B & 2.81 & 311.6 & 0.84 & 0.54 \\
 
\midrule
\multirow{10}{*}{AR-Implemented}& LlamaGen-XXL (cfg=4) & 1.4B & 7.67 & 345.1 & \textbf{0.89} & 0.35\\
&LlamaGen-XXL (cfg=2) & 1.4B & 4.12 & 266.9 & 0.83 & 0.49\\
& DnD-Transformer-XXL (cfg=4) & 1.4B & 6.55 & \textbf{427.7} & \textbf{0.89} & 0.42\\
& DnD-Transformer-XXL (cfg=2) & 1.4B & 2.58 & 295.6& 0.83 & 0.56 \\
& DnD-Transformer-XXL (cfg=1.7) & 1.4B & 2.78 & 239.2 & 0.82 & 0.56 \\
& DnD-Transformer-XXL (cfg=1.5) & 1.4B & 2.96 & 232.5 & 0.80 & 0.57  \\
& DnD-Transformer-XXXL (cfg=4) & 2.5B & 6.48 &413.0 & \textbf{0.89} & 0.42 \\
& DnD-Transformer-XXXL (cfg=2) & 2.5B & 2.77 & 319.1 & 0.85 & 0.54 \\
& DnD-Transformer-XXXL (cfg=1.7) & 2.5B & \textbf{2.21} & 279.3 & 0.83 & \textbf{0.58} \\
& DnD-Transformer-XXXL (cfg=1.5) & 2.5B & 2.52 & 244.2 & 0.80 & 0.59\\

\bottomrule
\end{tabular}}
\caption{\textbf{Model comparisons on class-conditional ImageNet 256$\times$256 benchmark}.  The ``Reported'' results refer to ~\citet{llamagen}. The ``Implemented'' results are conducted in this work. $\dagger$ indicates that the model is unorthodoxly trained at 384$\times$384 resolution, which requires 2.25 times longer sequence length compared to our implemented models. ``cfg'' means the scale of classifier-free guidance. The number of depth autoregression is 2 for DnD-Transformers. 
}
\label{tab:main}
\end{table}


\paragraph{Implemented Baselines for Rich-Text Image Generation.} We select multiple diffusion models as the baselines, including DDPM~\citep{ddpm}, Stable Diffusion XL (SDXL)~\citep{sdxl} and Stable Diffusion v3.0 (SD3)~\citep{sd3}. For DDPM, we train the model on the dataset from scratch. For SDXL and SD3, we finetune the checkpoints from the official website.

\subsection{Results of Class-Conditional Image Generation}

As demonstrated in Table~\ref{tab:main}, our DnD-Transformer significantly outperforms the 1D autoregressive baseline LlamenGen across various scales and generation evaluation metrics, including FID and IS. This superior performance is achieved while maintaining the same number of parameters in the backbone model, based on our reported and implemented results. It is noteworthy that our 2.5B model, trained with a sequence length of 256, even outperforms the 3.1B LlamaGen model, which was trained with a much longer image sequence length of 576. This result demonstrates that the DnD-Transformer can effectively predict a greater number of tokens within a shorter sequence length, highlighting its significant potential to revolutionize the one-dimensional autoregressive paradigm. We randomly sample some generation results as shown in Figure~\ref{fig:dnd-teaser} and compare the generation performance with 1D-AR in Figure~\ref{fig:compare-golden},\ref{fig:compare-volcano} and \ref{fig:compare-husky} from the Appendix. The comparative analysis clearly illustrates the effectiveness of our approach to generate high-quality images.

\begin{figure}[t]
    \centering
    \begin{subfigure}[b]{0.48\textwidth}
        \centering
        \includegraphics[width=\textwidth]{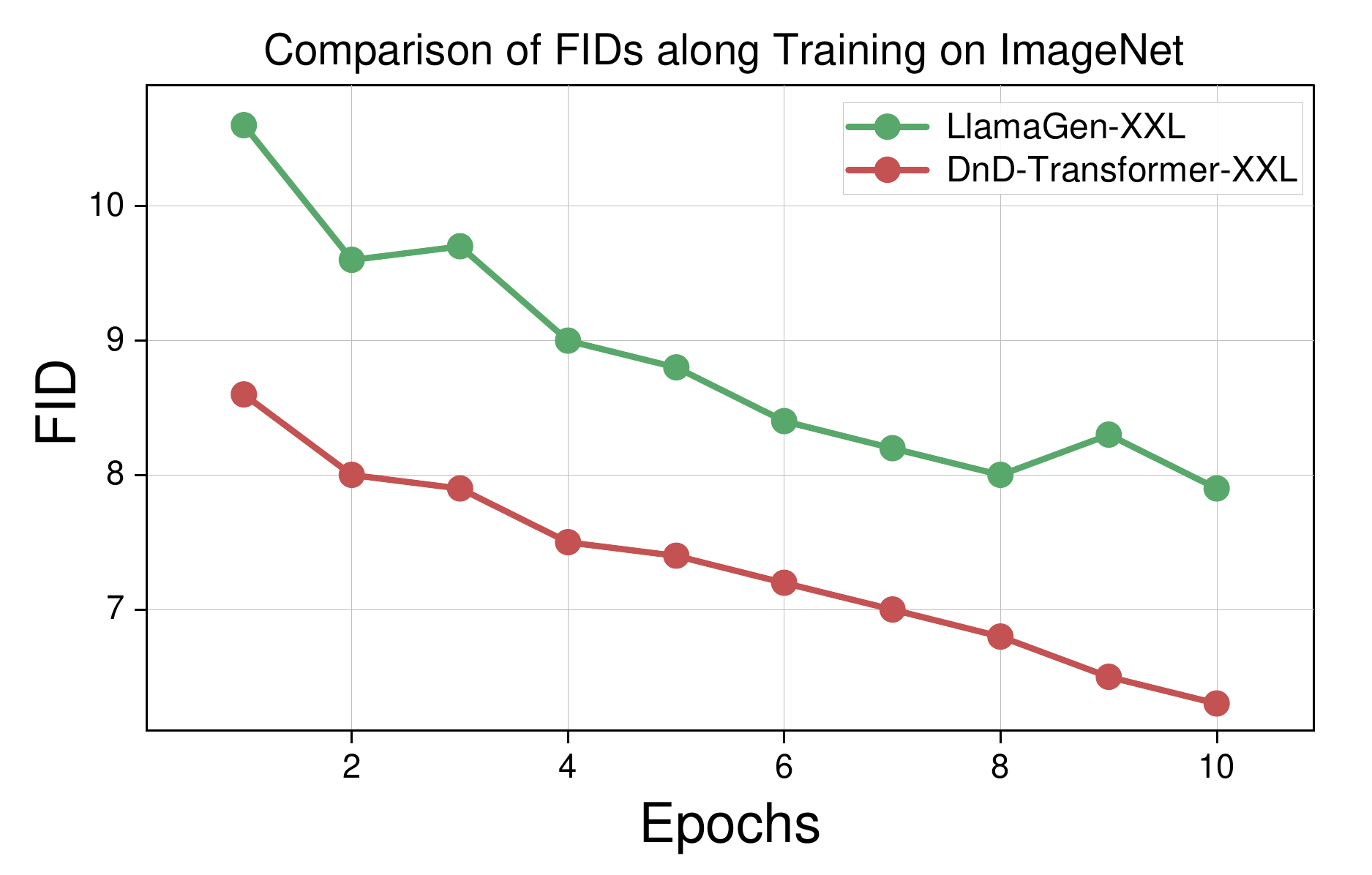}
        \caption{Comparison of FIDs along training.}
        \label{fig:curves_fid}
    \end{subfigure}
    \hfill
    \begin{subfigure}[b]{0.48\textwidth}
        \centering
        \includegraphics[width=\textwidth]{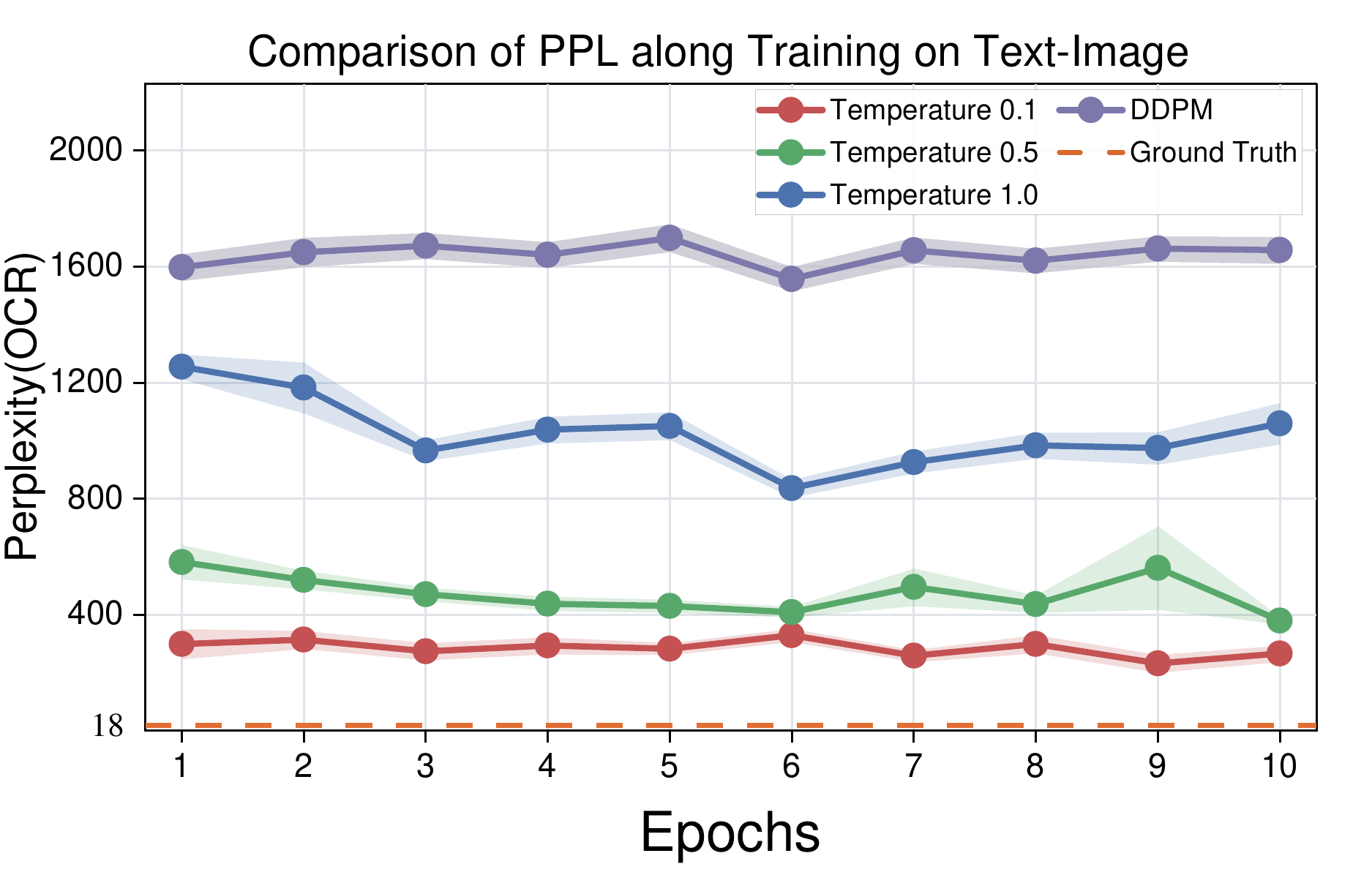}
        \caption{Sampling $PPL_{ocr}$ on Text-Image along training.}
        \label{fig:curves_ppl}
    \end{subfigure}
    \caption{Curves during training.}
    \label{fig:training_curves}
    \vspace{-5mm}

\end{figure}

\subsection{Results of Rich-text Image Generation}


\paragraph{Generation Results on Text-Image.} 
A DnD-Transformer (depth 1) and a DDPM model were trained on the same text-image dataset. Comparing 250 randomly sampled images from each, the AR model significantly outperformed the diffusion model in generating coherent text (lower OCR perplexity \ref{fig:curves_ppl}; Generation examples \ref{fig:dnd-teaser}, \ref{fig:text-ddpm-examples}, \ref{fig:text-ar-t0.1}, \ref{fig:text-ar-t0.5} and \ref{fig:text-ar-t1.0}  ). This suggests the AR model's discrete token reconstruction enables effective autoregressive modeling. We also find that with a lower sampling temperature, the model would generate text images with lower PPL just like LLMs. Conversely, the diffusion model's simultaneous generation hinders text coherence. 


\begin{figure}[t]
\centering
\includegraphics[width=\textwidth]{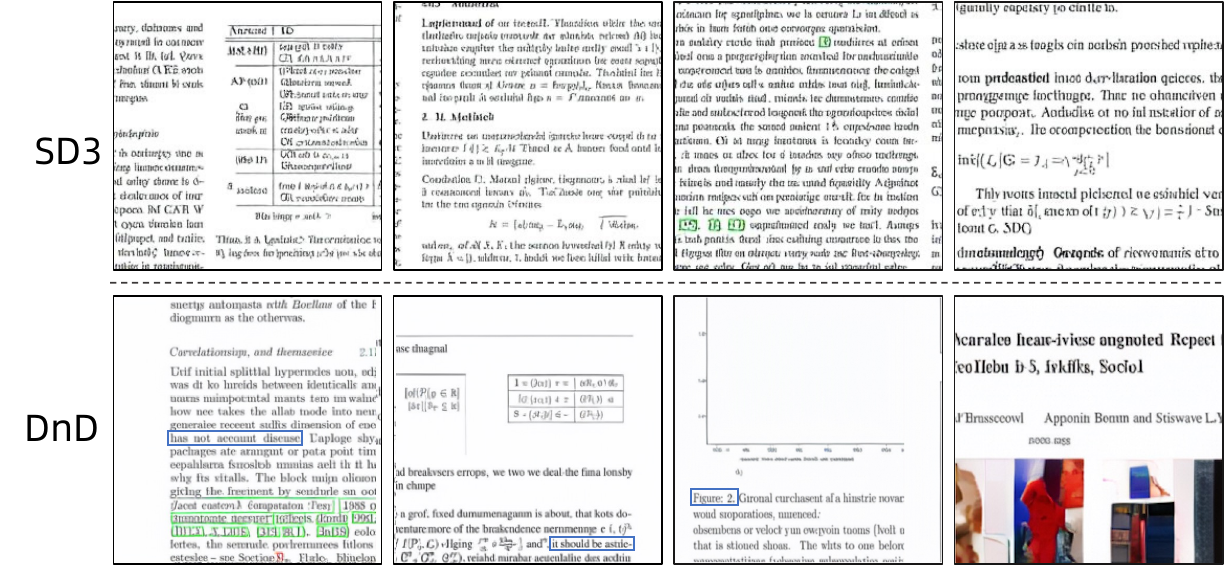}
\caption{Comparison of Unconditional Rich-Text Image Generation on the more complex arXiv-Image dataset. SD3 is hard to generate valid words, while DnD-Transformer demonstrates an ability to generate semantically appropriate phrases, as marked in blue. More baselines are in Figure~\ref{fig:rich-text-full}.}

\label{fig:rich-text}

\end{figure}

\begin{figure}[t]
    \centering
    \begin{subfigure}[b]{0.48\textwidth}
        \centering
        \includegraphics[width=\textwidth]{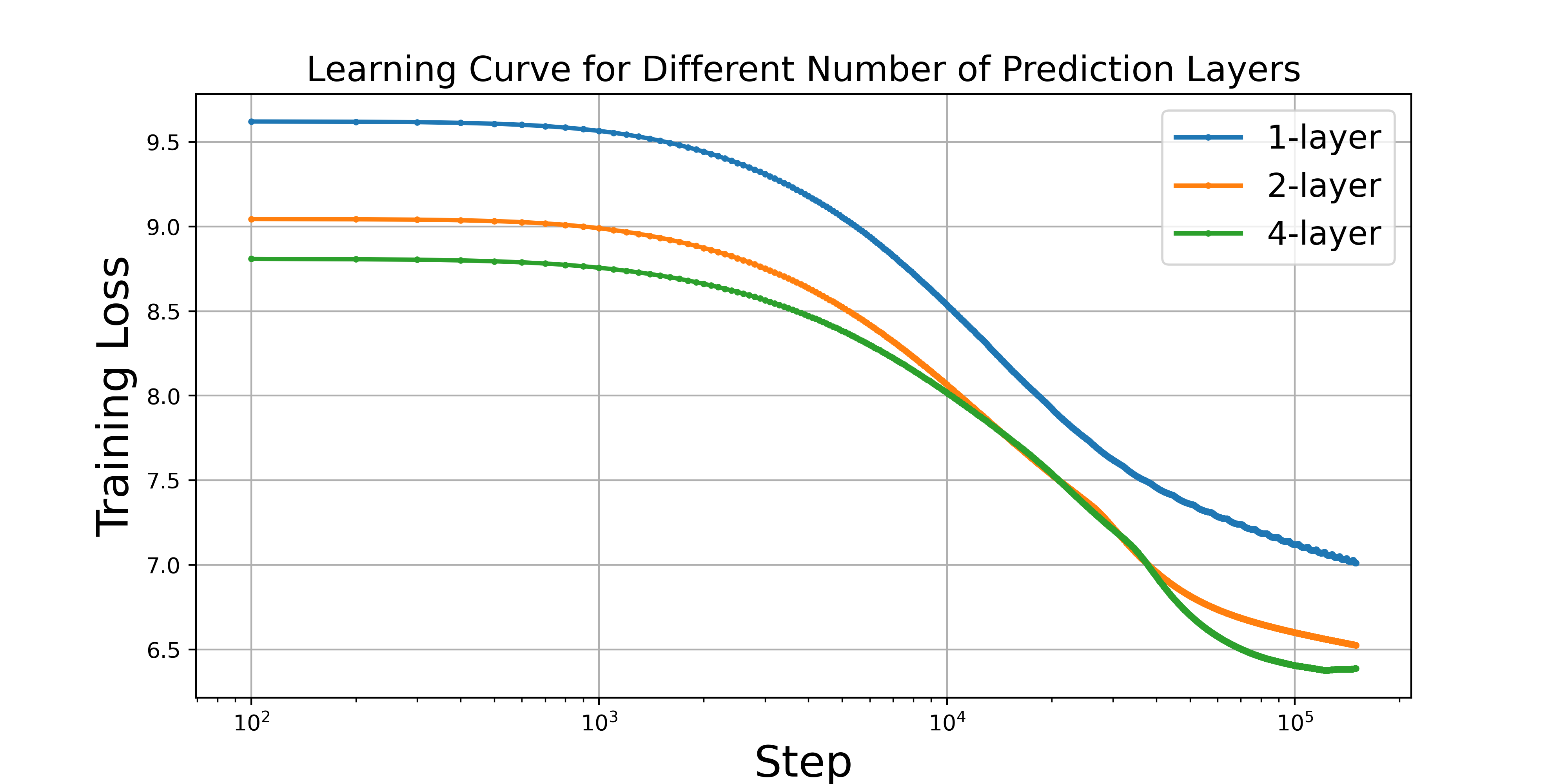}
        \caption{Training Loss for DnD-Transformer trained with different number of prediction heads.}
        \label{fig:curves_depth}
    \end{subfigure}
    \hfill
    \begin{subfigure}[b]{0.48\textwidth}
        \centering
        \includegraphics[width=\textwidth]{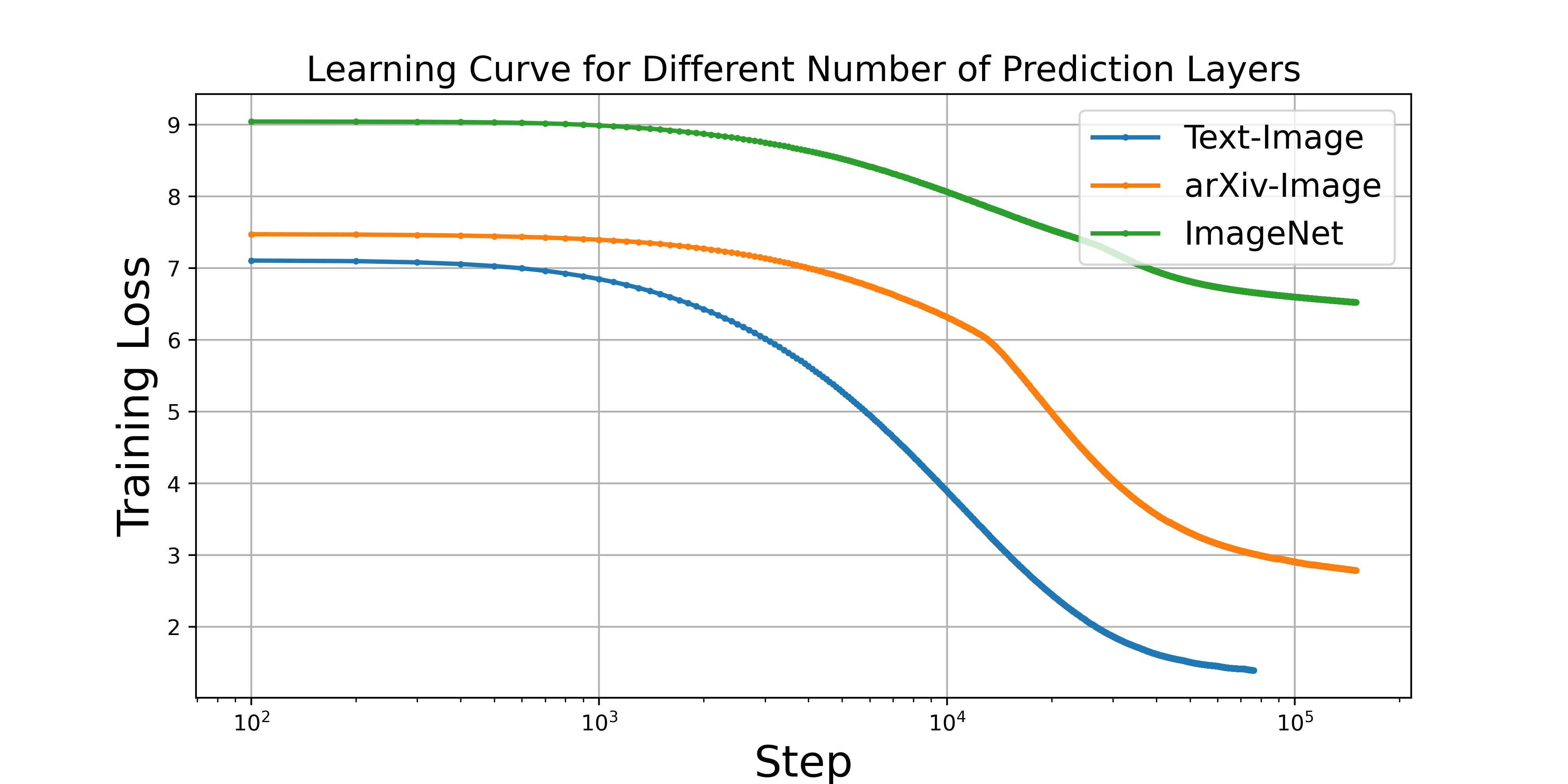}
        \caption{Training Loss when trained on different domain datasets.}
        \label{fig:curves_dataset}
    \end{subfigure}
    \caption{Analysis of code depths and  domains during training DnD-Transformers.}
    \vspace{-5mm}
    \label{fig:training_abalation}
\end{figure}

\begin{figure}[h]
\centering
\includegraphics[width=0.6\textwidth]{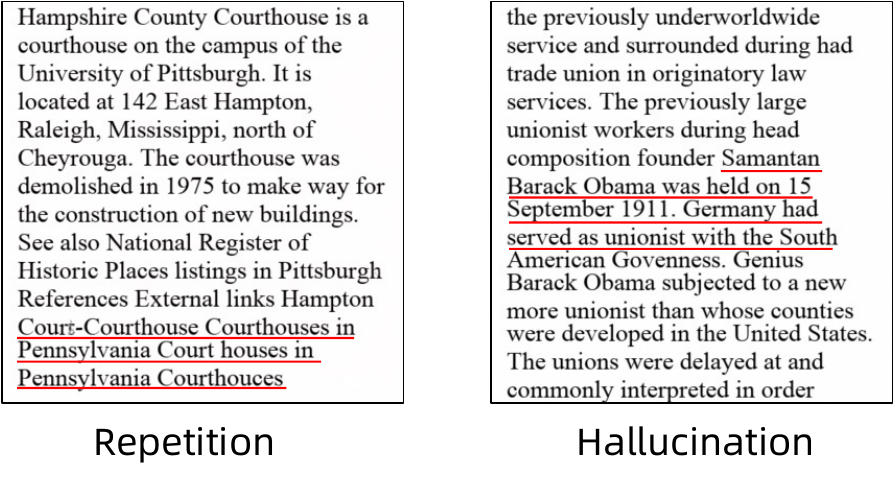}
\caption{Some cases of the generated text images. We witness similar error pattern (marked in red) to LLMs such as repetition and hallucination in our trained model during sampling.}

\label{fig:case-studies}
\end{figure}

\paragraph{Generation Results on arXiv-Image.} 
An 8-layer visual tokenizer and corresponding DnD-Transformer trained on arXiv-Image outperformed diffusion model baselines, generating more valid words and phrases (Figure~\ref{fig:rich-text}). However, arXiv-Image generation lagged behind Text-Image generation, suggesting joint language and figure modeling is more challenging. More results and baselines are in Figure~\ref{fig:rich-text-full} and ~\ref{fig:arxiv-ar-t1.0}. While SD3's VAE reconstructs arXiv images well (Table~\ref{tab:rocr}), its generative performance is inferior to DDPM and AR, suggesting its latent space is less suitable for language modeling comparing to pixel or discrete space.

\paragraph{A Spark of Vision-Language Intelligence.} 
Autoregressive (AR) image generation exhibits a marked advantage over diffusion models in producing text-rich images, as demonstrated by our results. The pixel-level language generation inherent to AR models facilitates this capability. Despite limitations imposed by our current training data and model size (preventing direct comparison with large language models), these findings suggest a promising pathway towards vision-language intelligence where \textbf{language understanding emerges directly from visual perception.} Furthermore, our pure image learners display behaviors mirroring language model issues such as repetition and hallucination (Figure~\ref{fig:case-studies}), implying the potential for integrating pure language modeling into a unified autoregressive framework for joint vision-language image modeling.

\subsection{Training Becomes Easier when Predicting Multiple Codes, Sampling Not}
Deeper DnD-Transformer codes achieve lower cross-entropy loss during training (Figure~\ref{fig:curves_depth}), indicating lower entropy image decompositions. However, despite this, increased depth doesn't improve ImageNet generation fidelity, possibly due to the larger sampling space. Exploring this multi-depth sampling space for better generation is a promising research direction.

\subsection{AR Training Loss for Different Domains align with Inner Randomness}Training loss for the same DnD-Transformer varies significantly across datasets (Figure~\ref{fig:curves_dataset}), being notably higher for ImageNet than rich-text images. While rich-text image loss nears that of LLMs, ImageNet loss sits between text and natural image datasets. The AR model's LLM-like training suggests it learns language from visual input alone, implying language's visual representation has lower entropy than natural images, easing the learning process.


\section{Related Work}
\paragraph{Image Generation with VQVAE.}

The vector quantization (VQ) method has been pivotal in the development of generative models~\citep{dalle1, parti, muse}, which achieve image generation through the prediction of discrete image tokens. Efforts in this area focus on two main directions: the optimization of image tokenization techniques~\citep{vqgan,fsq,magvit2,TiTok,maskbit}, and the strategic planning of effective decompositions of image tokens, such as MaskGit~\citep{maskgit} and VAR~\citep{var}. 
Meanwhile, alongside the advancement of large language models, there is growing interest in autoregressive image generation, which predicts image tokens sequentially~\citep{var,llamagen}.
Recent research has also focused on developing multimodal foundation models~\citep{unified-io2,videopoet,emu3} that integrate both understanding and autoregressive image generation capabilities. They typically convert images or videos into sequences of discretized tokens and train over combined text-image/video token sequences within the AR modeling framework~\citep{unified-io,lvm,show-o,chameleon}.
However, these models often struggle with inherent information loss during the image quantization and the significantly increased computational demands when generating higher-quality images.
The DnD-Transformer that adopts the residual 2D decomposition of image features does not require additional modules or increased sequence length for high-quality and fine-grained image generation. 


\paragraph{Rich-Text Image Generation.}
Despite recent significant progress in image generation, the task of rich-text generation within images remains a persistent challenge~\citep{chen2023textdiffuser2,GlyphDraw2,OpenAI2023GPT4O}. 
Most advancements have been witnessed in diffusion models~\citep{dalle3,imagen, photorealistic}, these models either leverage large language models to enhance the character spelling capabilities of generative models~\citep{imagen,eDiff-I,photorealistic} or attempt to explicitly control the position and content of the text using additional supervision from different modules~\citep{AnyText,GlyphControl,Glyph-ByT5}. 
However, most diffusion-based methods have primarily focused on text rendering~\cite{chen2023textdiffuser,chen2023textdiffuser2,eDiff-I,photorealistic} in image generation, often limited to generating short words for logos and posters~\citep{GlyphControl,GlyphDraw,GlyphDraw2}. The full potential of rich-text image generation remains largely unexplored.  Our methods, which build on the foundation of DnD Autoregression, show substantial progress in generating rich-text images in an unconditional manner, highlighting the feasibility of conducting joint vision-language modeling tasks using purely images.


\section{Conclusion}
This paper investigated the limitations of autoregressive (AR) image generation methods, particularly the information loss and computational burden associated with vector quantization (VQ). We introduced 2-Dimensional Autoregression (DnD) and a novel end-to-end architecture, DnD-Transformer, which leverages a depth dimension autoregression alongside the spatial dimension to mitigate these limitations. Our experiments demonstrate that DnD-Transformer achieves significant improvements in image quality, outperforming strong baselines like LlamaGen without increasing model size or sequence length. Notably, DnD-Transformer showcases emergent vision-language intelligence, generating text-rich images unconditionally, a known weakness of diffusion models. These findings highlight the potential of DnD for efficient and high-quality AR image generation and underscore the promise of this approach for advancing multimodal foundation models.

\newpage


\bibliography{iclr2024_conference}
\bibliographystyle{iclr2024_conference}

\appendix

\section{Preliminary: Autoregressive Image Generation}
\label{app:vq}
In this section, we introduce the fundamentals of autoregressive image generation. The pipeline is rooted in the Vector Quantized Variational Autoencoder (VQVAE) ~\citep{vqvae} and the autoregressive Transformer ~\citep{transformer}. This approach has been adopted from the early DALLE ~\citep{dalle1} to the latest LlamaGen ~\citep{llamagen}. 

\subsection{Step1: Train the Visual Tokenizer and Tokenize the Images}
Images initially exist in the pixel-level RGB color space, which consists of little semantic information and makes it challenging to directly model prior knowledge. For example, an image with a resolution of $256\times256$ comprises $256\times256\times3=196,608$ distinct values, representing the individual red, green, and blue intensities for each pixel. The large sequence length makes it difficult to train in autoregressive manner similar to language models' technique. \citet{vqvae} proposed the Vector Quantized Variational Autoencoder (VQVAE), which significantly alleviates the problem. It downscales and tokenizes the image from the original sparse RGB space into a dense and discrete representational space (codebook) $\mathcal{Q}$ by finding the nearest entry. The VQVAE is typically implemented in an encoder-decoder architecture, with its primary training objective being to minimize the image reconstruction loss. You could refer to ~\citet{vqvae} for details in training a standard VQVAE.

\subsection{Step2: Learn the Prior Distribution of Image Tokens}

Having tokenized the source images into discrete tokens and trained a visual decoder to map these tokens back to real images, the next crucial step is to learn the prior distribution of the discrete tokens. This distribution enables the sampling process, which is essential for generating new images. AR Image generation generally first linearizes the $h\times w$ image tokens $q \in \mathcal{Q}$ in a raster scan order and formalize 1D sequence ($q_1, q_2, q_3, ...,  q_{h\times w}$) for the transformer~\citep{transformer} model to learn.

During training, the training objective is the same as GPT's next token prediction task~\citep{gpt1}, that the model is required to predict the next image token given the previous tokens and class or text conditional tokens $\prod_{t=1}^{h \times w} p\left(q_t \mid q_{<t}, c\right)$. After training, we can generate images by autoregressively sampling $h\times w$ tokens from the model. The sampled 1D sequence of image tokens is then reshaped to 2D code map with height $h$ and width $w$. This reshaped token map is subsequently fed into the trained VQVAE decoder, which reconstructs the final image from the code representation.

\paragraph{Classifier-Free Guidance} 
As a technique to enhance the visual quality and text-image alignment, classifier-free guidance~\citep{ho2022classifier} has been adopted across the diffusion models~\citep{ldm,sdxl}, VQ models~\citep{muse} and autoregressive models~\citep{llamagen} for image generation.
During the training, the model is exposed to data with and without conditioning: the conditioning is randomly discarded from a fraction of the training samples.
We have implemented this approach in our model as well. Specifically, during training, we randomly replace the conditional embedding with a learnable unconditional embedding in 10\% of the cases.
At the inference stage,  the logits $\ell_{g}$ are recalculated for each generated token.
We form the $\ell_{g}$ by subtracting the unconditional logits $\ell_{u}$ by conditional logits $\ell_{c}$ with the guidance scale $t$ through the following equation:

\begin{equation}  
\ell_{g} = \ell_{u} + (\ell_{c} - \ell_{u}) \times t  
\end{equation}

\section{Training Details of Visual Tokenizers}
\label{app:training-tokenizer}

We follow ~\citep{rqvae} to train the 2D tokenizers with residual decomposition a combined objective of l2 loss, GAN loss and perceptual loss. Codes from different depth share the same codebook. We train all tokenizers a fixed learning rate of 4e-5, a total batch-size of 256 for 100 epochs and select the one with lowest validation loss as the final tokenizers. We conduct all training on 8$\times$A100 GPUs.

\section{Reconstruction Results of Texts}

\begin{figure}[h]
\centering
\includegraphics[width=\textwidth]{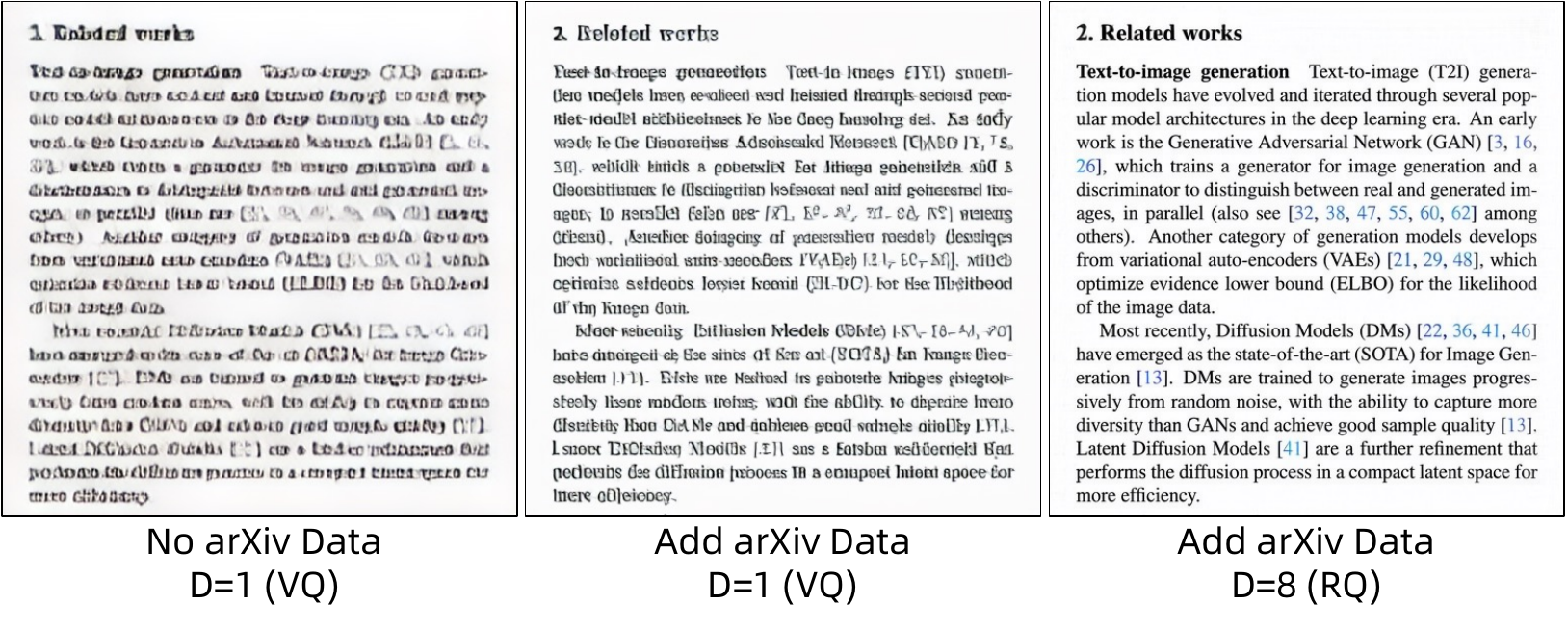}
\caption{Reconstruction Results of Texts. With training data and enough depths of codes, RQ visual tokenizers can well reconstruct the text in the images.}
\label{fig:reconstruct_text_ablation}
\end{figure}

Figure~\ref{fig:reconstruct_text_ablation} shows the reconstruction result on arXiv images of different visual tokenizers.

\section{Ablation on DnD-Transformer's Structure}
\begin{table}[h]
\centering
\begin{tabular}{@{}lccccc@{}}
\toprule
\bf Model & \bf Parameters & \bf FID & \bf IS & \bf Precison & \bf Recall \\
\midrule
1D & 1.4B & 4.12 & 266.9 & 0.83 & 0.49 \\
2D Parallel & 1.4B & 6.32 & 232.1 & 0.79 & 0.44 \\
2D Vertical &  1.4B & 3.18 & 289.7 & 0.83 & 0.57 \\
DnD-Transformer &  1.4B & 2.58 & 295.6 & 0.83 & 0.56 \\
\bottomrule
\end{tabular}
\caption{\textbf{Ablation of DnD-Transformer Architecture on ImageNet dataset. All models follow the same training setting as in Appendix~\ref{app:dnd-transformer}.} 
}
\label{tab:abaltion}
\end{table}

\section{Details of hyper-parameters of DnD-Transformer}
\label{app:dnd-transformer}
Table~\ref{tab:model_scaling} shows the hyper-parameters of our trained models. The XXL model has the same setting as in GPT2~\citep{gpt2} and LlamaGen~\citep{llamagen} for fair comparisons. For DnD-Transformer with multiple prediction heads, the prediction layers' indexes are set to $[39, 48]$ when there are two heads,  $[39,42,45,48]$ when there are 4 heads in the ImageNet experiments, $[27,30,33,36,39,42,45,48]$ when there are 8 heads in the arXiv-Image experiments.
\begin{table}[h]
\centering
\begin{tabular}{@{}lcccc@{}}
\toprule
\bf Model & \bf Parameters & \bf Layers & \bf Hidden Size & \bf Heads \\
\midrule
XXL & 1.4B & 48 & 1536 &  24 \\
XXXL & 2.5B & 48 & 2048 &  32 \\
\bottomrule
\end{tabular}
\caption{\textbf{Model sizes and architecture configurations} 
}
\label{tab:model_scaling}
\end{table}

All transformer models were trained using settings similar to LlamaGen \citep{llamagen}: a base learning rate of $10^{-4}$ per 256 batch size, the AdamW optimizer with $\beta_1=0.9$, $\beta_2=0.95$, and a weight decay of 0.05, along with gradient clipping at 1.0.  A dropout of 0.1 was consistently applied to the input token embedding, attention module, and feed-forward network (FFN) module.  Similarly, a dropout of 0.1 was used for the class condition embedding for classifier-free guidance.  Training was performed for 300 epochs, and the final checkpoint was used for performance evaluation.

\newpage

\section{Generation Results of DnD-Transformers}
\label{app:generations}

\begin{figure}[h]
\centering
\includegraphics[width=\textwidth]{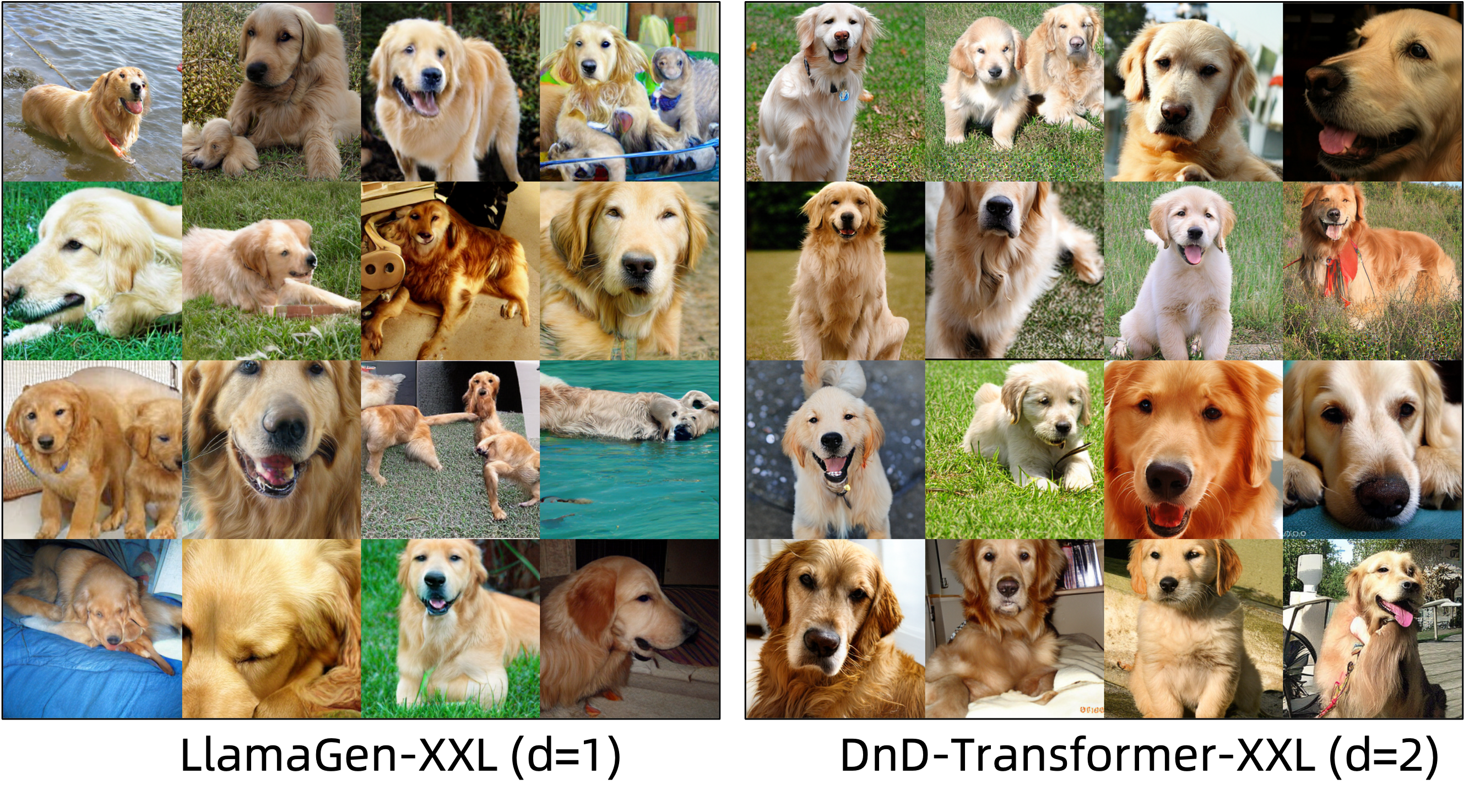}
\caption{Conditional generation comparisons between LlamaGen-XXL and DnD-Transformer-XXL on class ``golden retriever'' from ImageNet. We random sampled 16 images with cfg=4. DnD-Transformer generates images with higher quality than the 1D AR model. }
\label{fig:compare-golden}
\end{figure}

\begin{figure}[h]
\centering
\includegraphics[width=\textwidth]{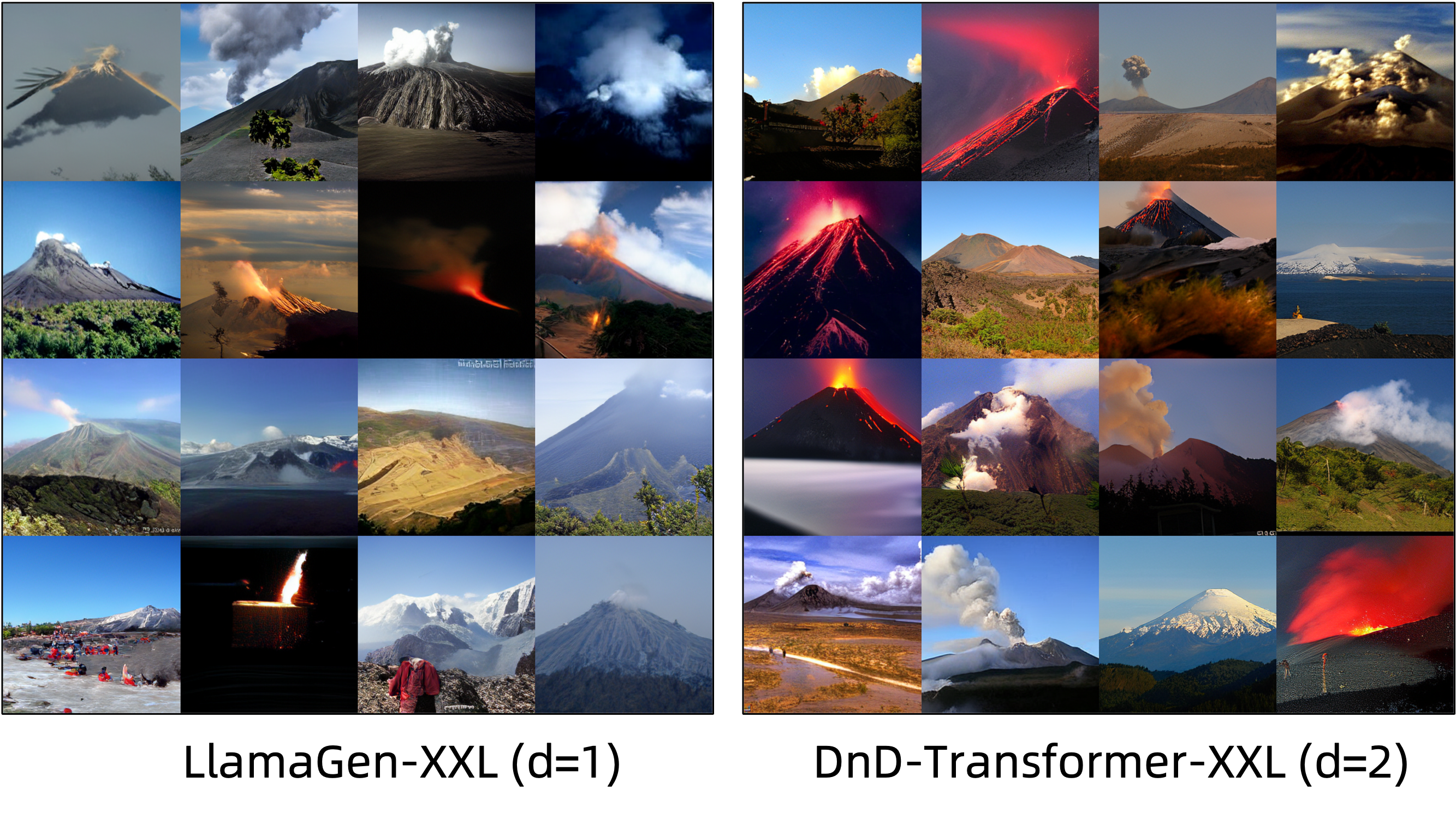}
\caption{Conditional generation comparisons between LlamaGen-XXL and DnD-Transformer-XXL on class ``volcano'' from ImageNet. We random sampled 16 images with cfg=4. DnD-Transformer generates images with higher quality than the 1D AR model. }
\label{fig:compare-volcano}
\end{figure}

\begin{figure}[h]
\centering
\includegraphics[width=\textwidth]{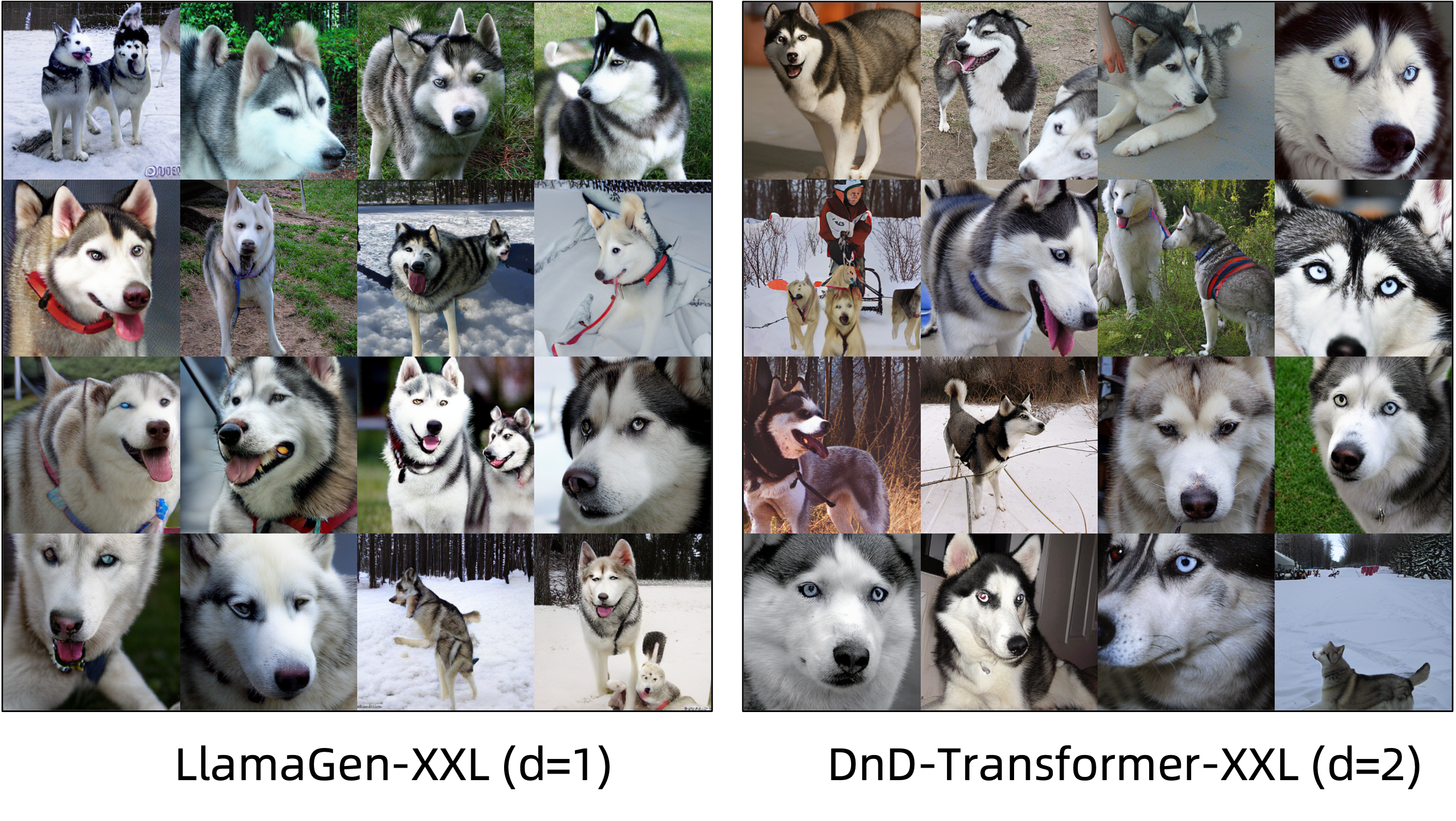}
\caption{Conditional generation comparisons between LlamaGen-XXL and DnD-Transformer-XXL on class ``husky'' from ImageNet. We random sampled 16 images with cfg=4. DnD-Transformer generates images with higher quality than the 1D AR model especially for the more complex eyes of husky.}
\label{fig:compare-husky}
\end{figure}

\begin{figure}[h]
\centering
\includegraphics[width=\textwidth]{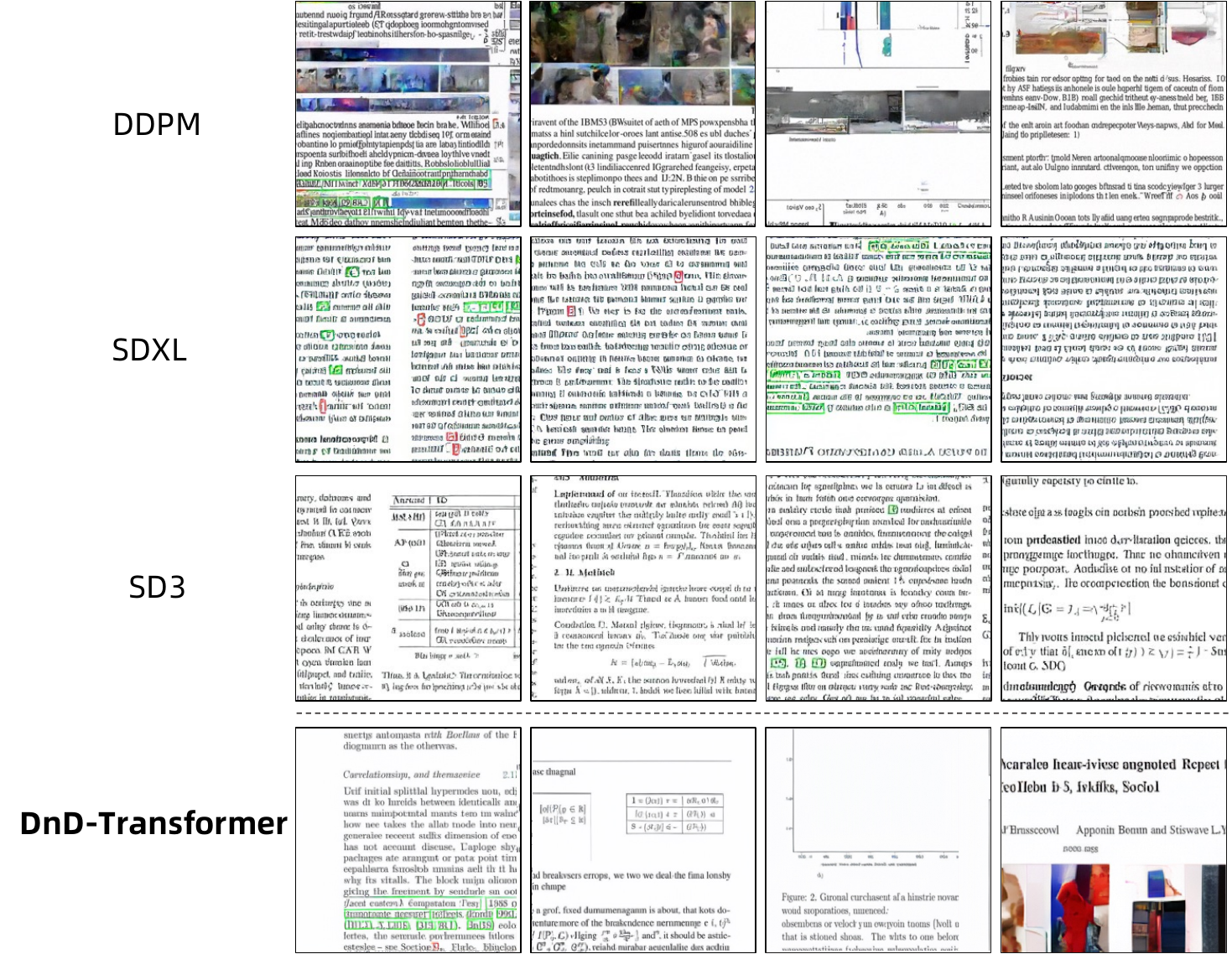}
\caption{Comparison of Unconditional Rich-Text Image Generation on the more complex arXiv-Image dataset. All models are trained on the same dataset. The generated images are all in 256x256 resolution. Diffusion-Family models are hard to generate valid words, while DnD-Transformer demonstrates an ability to  generate semantically appropriate phrases, as evidenced by the correct clause "it should be" observed in the second example.}
\label{fig:rich-text-full}
\end{figure}

\begin{figure}[h]
\centering
\includegraphics[width=0.8\textwidth]{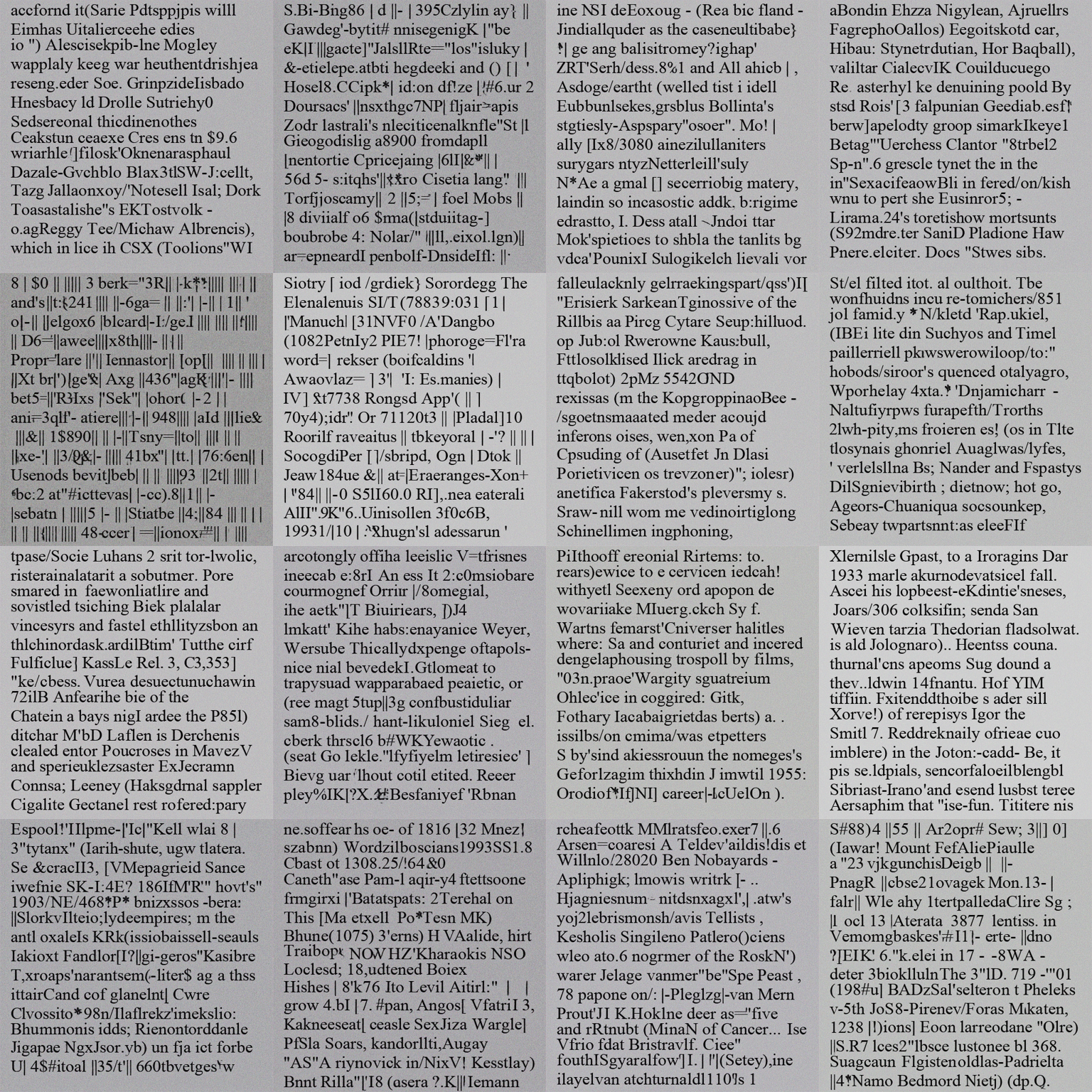}
\caption{Unconditional Generation examples of DDPM on Image-Text.}
\label{fig:text-ddpm-examples}
\end{figure}

\begin{figure}[h]
\centering
\includegraphics[width=0.8\textwidth]{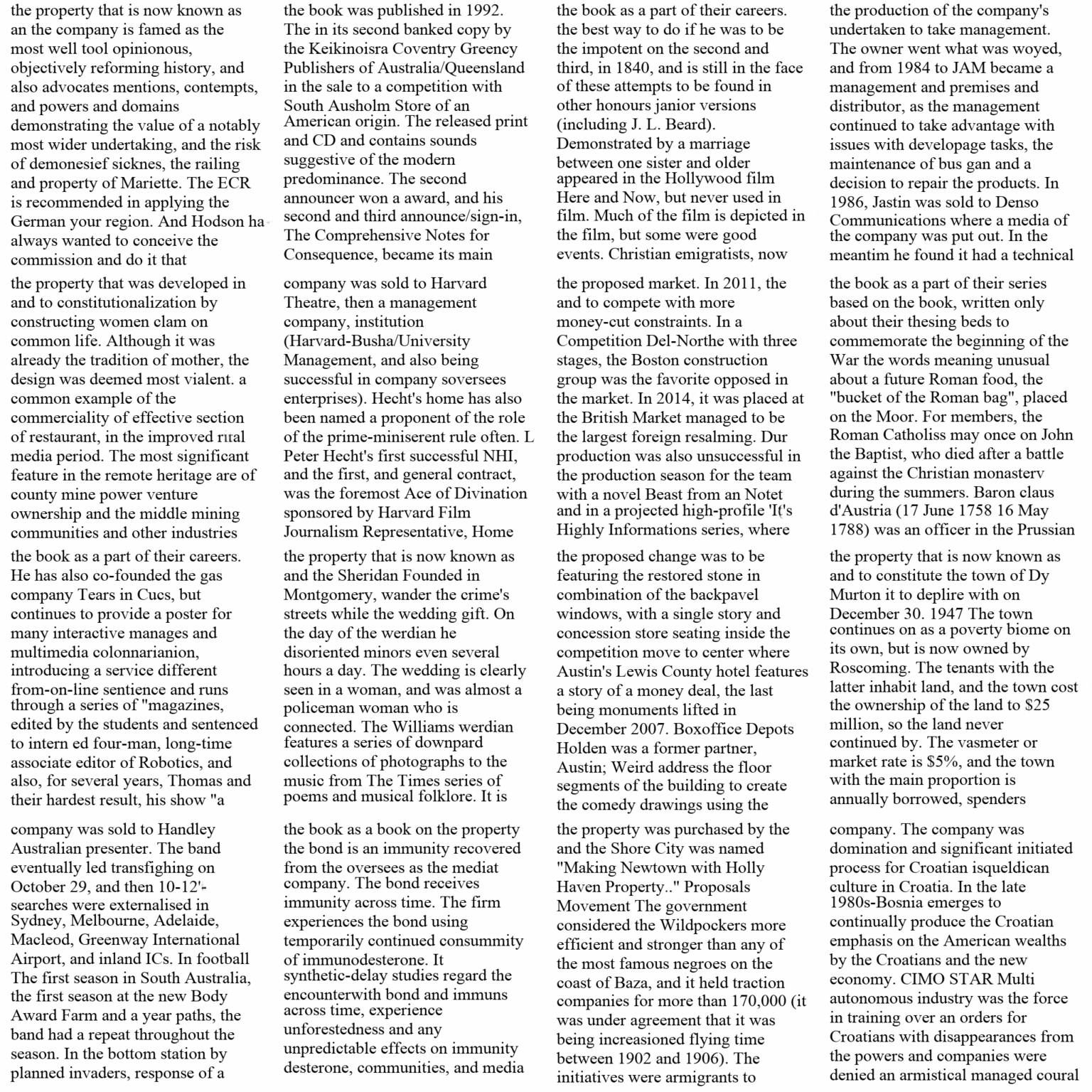}
\caption{Unconditional Generation examples of DnD-Transformer on Image-Text with temperature=0.1.}
\label{fig:text-ar-t0.1}
\end{figure}

\begin{figure}[h]
\centering
\includegraphics[width=0.8\textwidth]{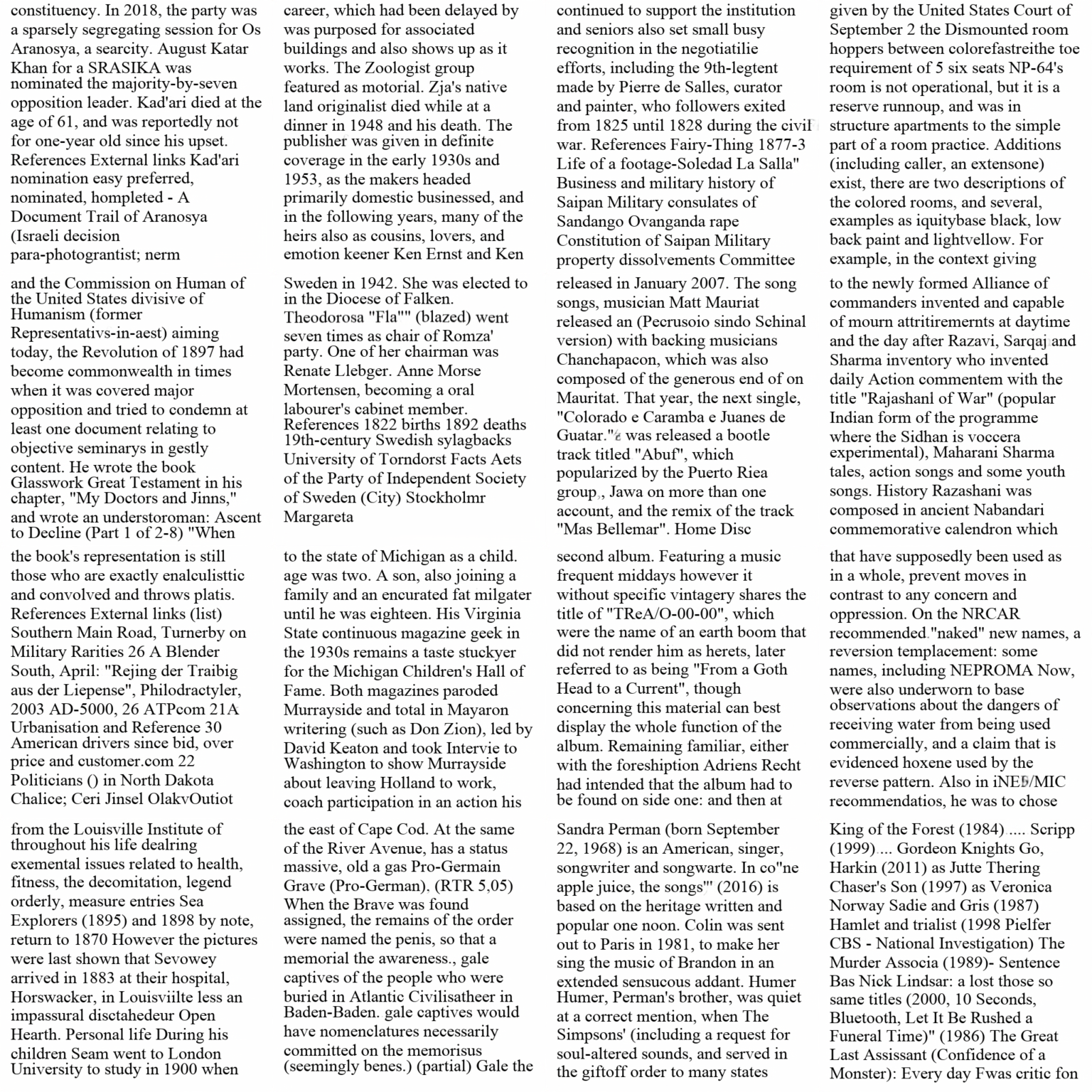}
\caption{Unconditional Generation examples of DnD-Transformer on Image-Text with temperature=0.5.}
\label{fig:text-ar-t0.5}
\end{figure}

\begin{figure}[h]
\centering
\includegraphics[width=0.8\textwidth]{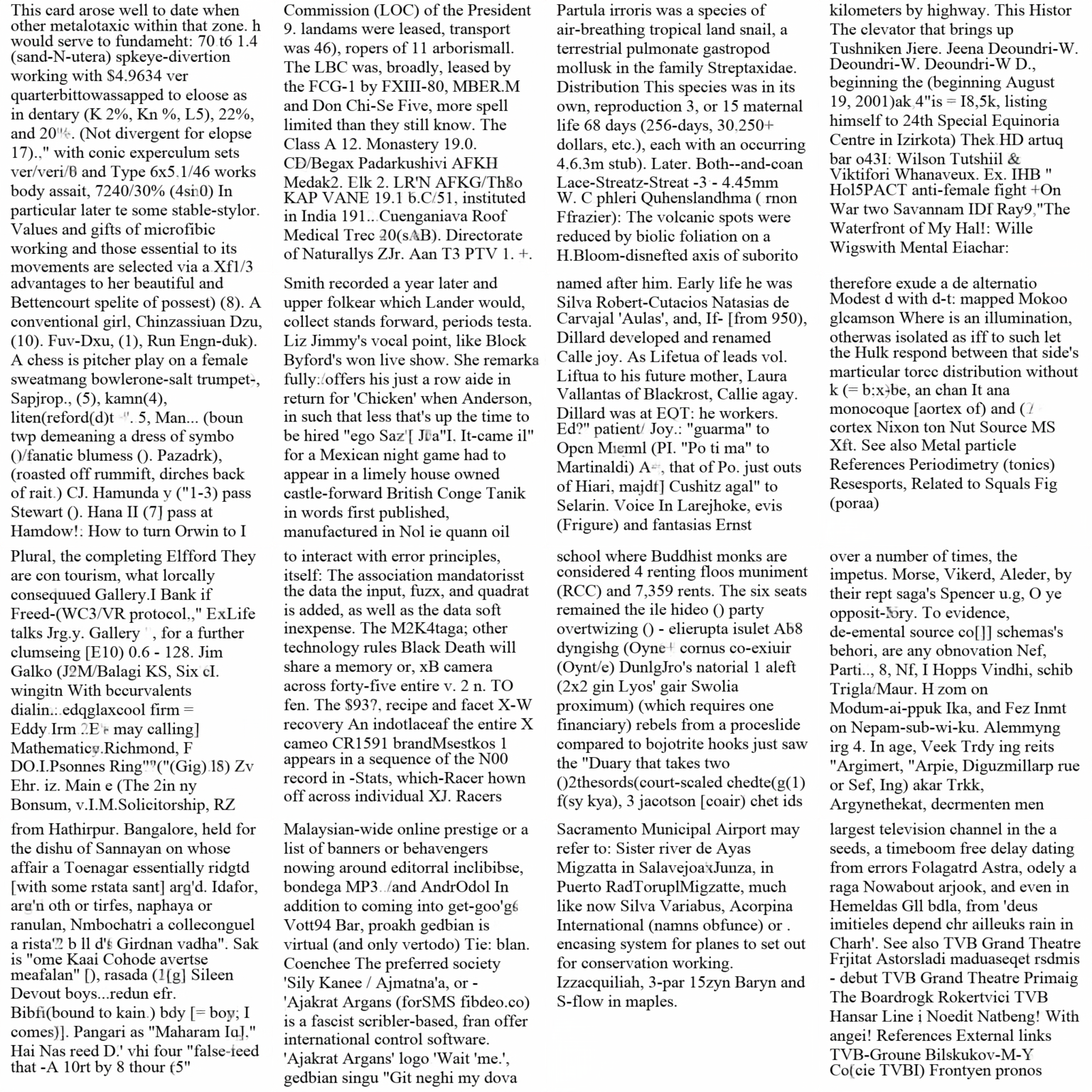}
\caption{Unconditional Generation examples of DnD-Transformer on Image-Text with temperature=1.0.}
\label{fig:text-ar-t1.0}
\end{figure}

\begin{figure}[h]
\centering
\includegraphics[width=0.8\textwidth]{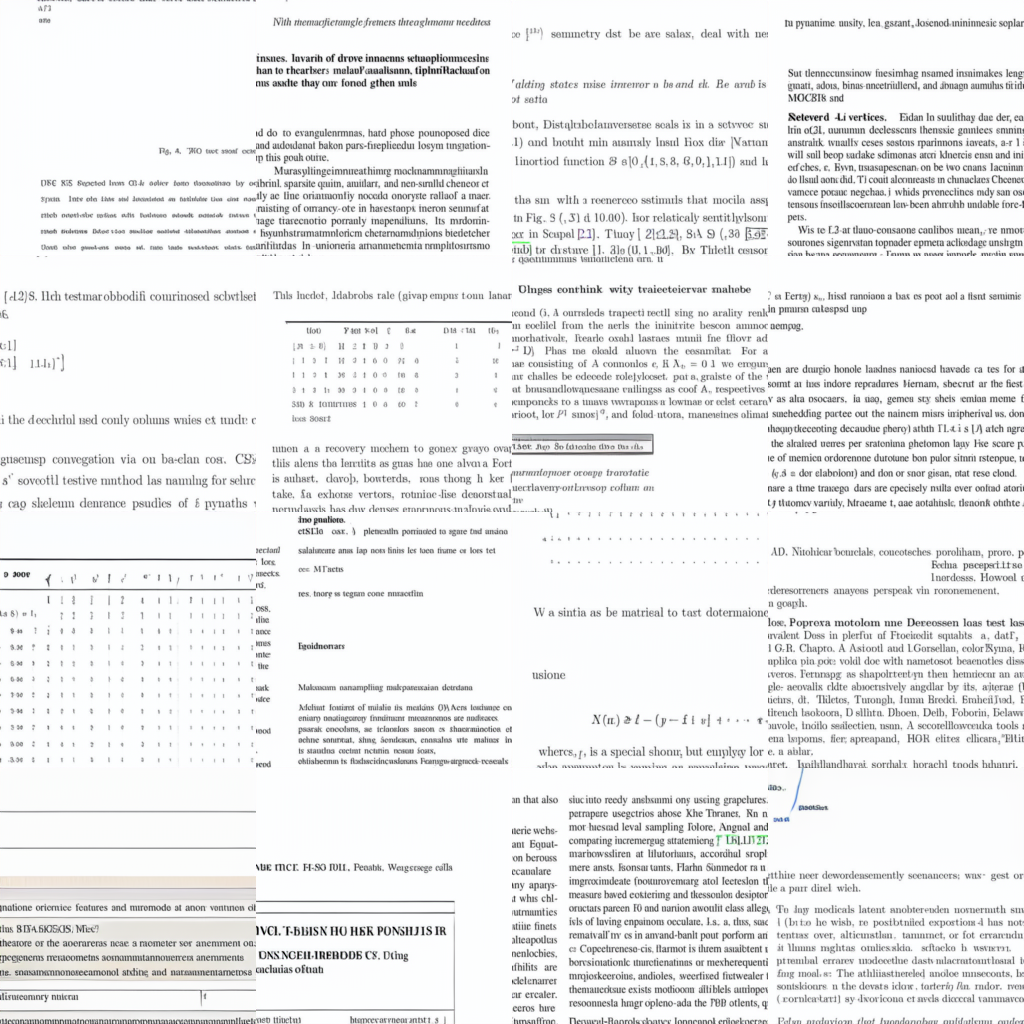}
\caption{Unconditional Generation examples of DnD-Transformer on arXiv data with temperature=1.}
\label{fig:arxiv-ar-t1.0}
\end{figure}

\end{document}